\RequirePackage{fix-cm}
\documentclass[smallcondensed]{svjour3}     
\smartqed  
\usepackage{graphicx}
\usepackage{wrapfig}
\usepackage{hyperref}
\usepackage[bottom]{footmisc} 
\usepackage{soul}
\usepackage{xcolor}

\usepackage{xcolor,colortbl}
\definecolor{Gray}{gray}{0.25}
\definecolor{red}{rgb}{1.00,0.00,0.00}
\definecolor{blue}{rgb}{0.00,0.00,1.00}
\definecolor{green}{rgb}{0.2,0.50,0.0}
\definecolor{yellow}{rgb}{0.5,0.5,0.0}

\begin{document}

\title{The State of {Lifelong Learning in} Service Robots:}
\subtitle{Current Bottlenecks in Object Perception and Manipulation}


\author{S. Hamidreza Kasaei \and
        Jorik Melsen \and 
        Floris van Beers \and 
        Christiaan Steenkist \and 
        Klemen Voncina
}


\institute{All authors are with Department of Artificial Intelligence, University of Groningen, PO Box 407, 9700 AK, Groningen, The Netherlands.
\email {hamidreza.kasaei@rug.nl} , \{j.l.a.melsen, f.van.beers, c.n.steenkist, k.voncina\}@student.rug.nl }

\date{Received: date / Accepted: date}

\maketitle

\begin{abstract}
Service robots are appearing more and more in our daily life. The development of service robots combines multiple fields of research, from object perception to object manipulation. The state-of-the-art continues to improve to make a proper coupling between object perception and manipulation. This coupling is necessary for service robots not only to perform various tasks in a reasonable amount of time but also to continually adapt to new environments and safely interact with non-expert human users. Nowadays, robots are able to recognize various objects, and quickly plan a collision-free trajectory to grasp a target object {in predefined settings.}
Besides, in most of the cases, there is a reliance on large amounts of training data. Therefore, the knowledge of such robots is fixed after the training phase, and any changes in the environment require complicated, time-consuming, and expensive robot re-programming by human experts. Therefore, these approaches are still too rigid for real-life applications in unstructured environments, where a significant portion of the environment is unknown and cannot be directly sensed or controlled. {In such environments, no matter how extensive the training data used for batch learning, a robot will always face new objects. Therefore, apart from batch learning, the robot should be able to continually learn about new object categories and grasp affordances from very few training examples on-site. Moreover, apart from robot self-learning, non-expert users could interactively guide the process of experience acquisition by teaching new concepts, or by correcting insufficient or erroneous concepts. In this way, the robot will constantly learn how to help humans in everyday tasks
by gaining more and more experiences without the need for re-programming.} In this paper, we review advances in service robots from object perception to complex object manipulation and shed light on the current challenges and bottlenecks.

\end{abstract}

\begin{figure}[!b]
\centering
\includegraphics[width=0.9\linewidth]{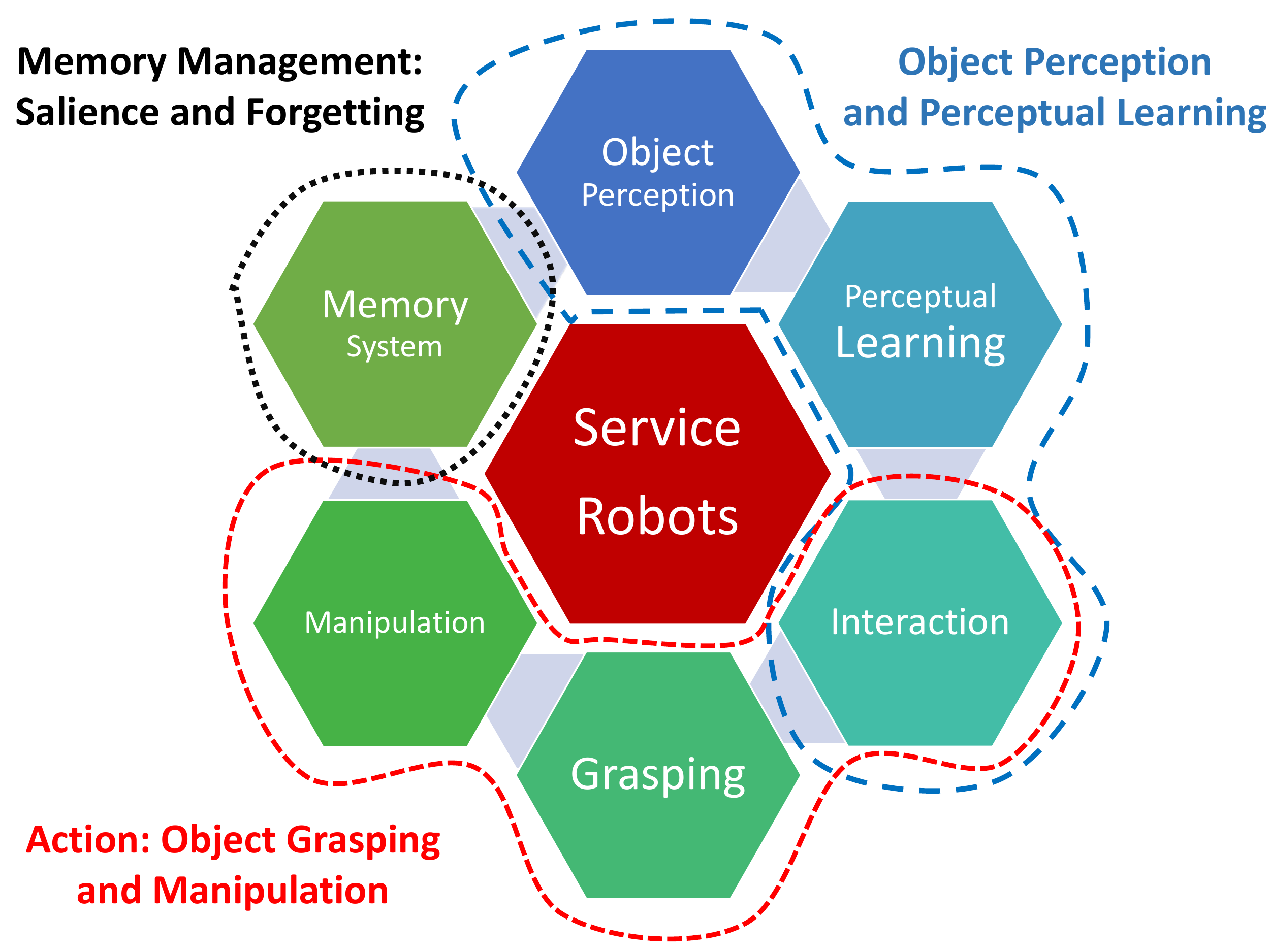}
\caption{ Categorization of {object-related} sub-tasks of a service robots into three core components, including: (\textit{i}) object perception and perceptual learning, (\textit{ii}) object grasping and manipulation, (\textit{iii}) memory management. }
\label{required_modules}
\end{figure}

\section{Introduction}
The development of service robots, used for any domestic or service task, is ongoing and interest is growing. On the one hand, there is an increase in supply, with ever more efficient and widely applicable robots. On the other hand, there is an increase in demand, due to interest from service industries to automate as well as an increasing elderly population~\cite{spasova2018challenges} that is a significant challenge for many countries. 
According to a recent study~\cite{spasova2018challenges}, the number of Europeans aged 80-plus is set to rise from 4.9\% in 2016 to 13\% in 2070. The old-age dependency ratio of the European
population (i.e., people aged 65 or above relative to those aged 15-64) was 29.6\% in 2016 and is projected to reach 51.2\% by 2070, meaning for every person in retirement age there are less than two people of working age. This significant demographic change poses several challenges since the population of caregivers is shrinking, while the number of people needing care is growing. This imbalance between demand and supply, leads to a big gap in the workforce and, therefore, calls for developing service robots to help us overcome this ever-increasing problem.

There is a wide diversity of tasks that a service robot can be used for, such as setting a table for a meal, clearing a table after eating a meal, serving a drink, helping people carry groceries \cite{robocup2018}, etc. Most of these household tasks can be decomposed into detecting an object, driving the robot's arm to a desired pose, and manipulating an object.
In other words, a robot needs to know which kinds of objects exist in a scene, where they are, and how to grasp and manipulate objects in different situations to operate in human-centric domains. These tasks are of a high complexity and consist of several sub-tasks that need to be performed sequentially or simultaneously to accomplish a certain goal.
As shown in Fig.~\ref{required_modules}, we have categorized these sub-tasks into three core components, including:

\begin{itemize}
    \item \textbf{Object perception and perceptual learning:} A service robot may sense the world through different modalities. The perception system provides important information that the robot has to use for interacting with users and environments. For instance, to interact with users and environment, a robot needs to know which kinds of objects exist in a scene and where they are. Besides, learning mechanisms allow incremental and open-ended learning. {A service robot must update its models over time by interacting with non-expert users and with limited computational resources {or access to cloud services}. Interaction and communication is one of the effective {ways} for a robot to obtain knowledge from a human user/teacher. Therefore, an interaction interface that facilitates language transfer from a human user to the robotic agent is an important module that a cognitive architecture should support. Interaction capabilities are mainly developed to enable human users to interactively guide the process of experience acquisition by teaching new concepts or by correcting insufficient or erroneous concepts, and instruct the robot to grasp objects and perform complex manipulation tasks~\cite{langley2009cognitive}}.

    \item \textbf{Object grasping and manipulation:}  A service robot must be able to grasp and manipulate objects in different situations to interact with the environment as well as human users. {Towards this goal, the robot should be able to {learn how to grasp objects}, generate plans and solve problems to achieve adaptability. Furthermore, it must be able to execute skills and actions in the environment. In most of the frameworks, object manipulation mainly works in a completely reactive manner. In other words, the agent selects one or more primitive actions on each decision cycle, executes them, and repeats the process on the next cycle. This approach is associated with closed-loop strategies for execution since the agent can also sense the environment on each time step. It should be noted that \textit{Task Planning} is not in the scope of this paper, and we refer the reader to the relevant surveys of Ingrand et al.,~\cite{ingrand2017deliberation} and Tsarouchi et al.,~\cite{tsarouchi2016human} for more details on task planning.}

    \item \textbf{Memory management:} A service robot, working in an open-ended domain, should involve experience management mechanisms such as salience and forgetting to prevent the accumulation of examples in the memory. Otherwise, the memory consumption and the required time to both update the models and recognize new objects would increase exponentially. 
\end{itemize}

\noindent
These core components are tightly coupled together using a software architecture. This coupling is necessary for service robots, not only to perform object perception and manipulation tasks in a reasonable amount of time, but also to robustly adapt to new environments by handling new objects. {In \textit{mobile} service robots, simultaneous localization and mapping (SLAM) is as crucial as the other three mentioned modules.  In particular, a mobile robot must determine its location within the environment and navigate safely in the environment. In this paper, we mainly focus on object perception and object manipulation and do not cover the topic of SLAM in mobile robots. We refer the reader to recently published review papers for more details on mobile robot navigation and localization~\cite{zhao2019compatible,yasuda2020autonomous,debeunne2020review}.}

A service robot should process very different types of information in varying time scales. Two different modes of processing, generally labelled as System 1 and System 2, are commonly accepted theories in cognitive psychology~\cite{evans2008dual}, and are mainly used as a software architecture of a service robot. The operations of System 1 (i.e., perception and action) are typically fast, automatic, reactive, and intuitive. The operations of System 2 (i.e. semantic) are slow, deliberative, and analytic. In this review paper, we mainly focus on object perception and manipulation tasks in service robots with the distinctive characteristics of System 1. 
\begin{wrapfigure}{r}{0.5\linewidth}
    \vspace{-2mm}
    \includegraphics[width=\linewidth,trim= 0cm 0.5cm 20cm 0cm,clip=true]{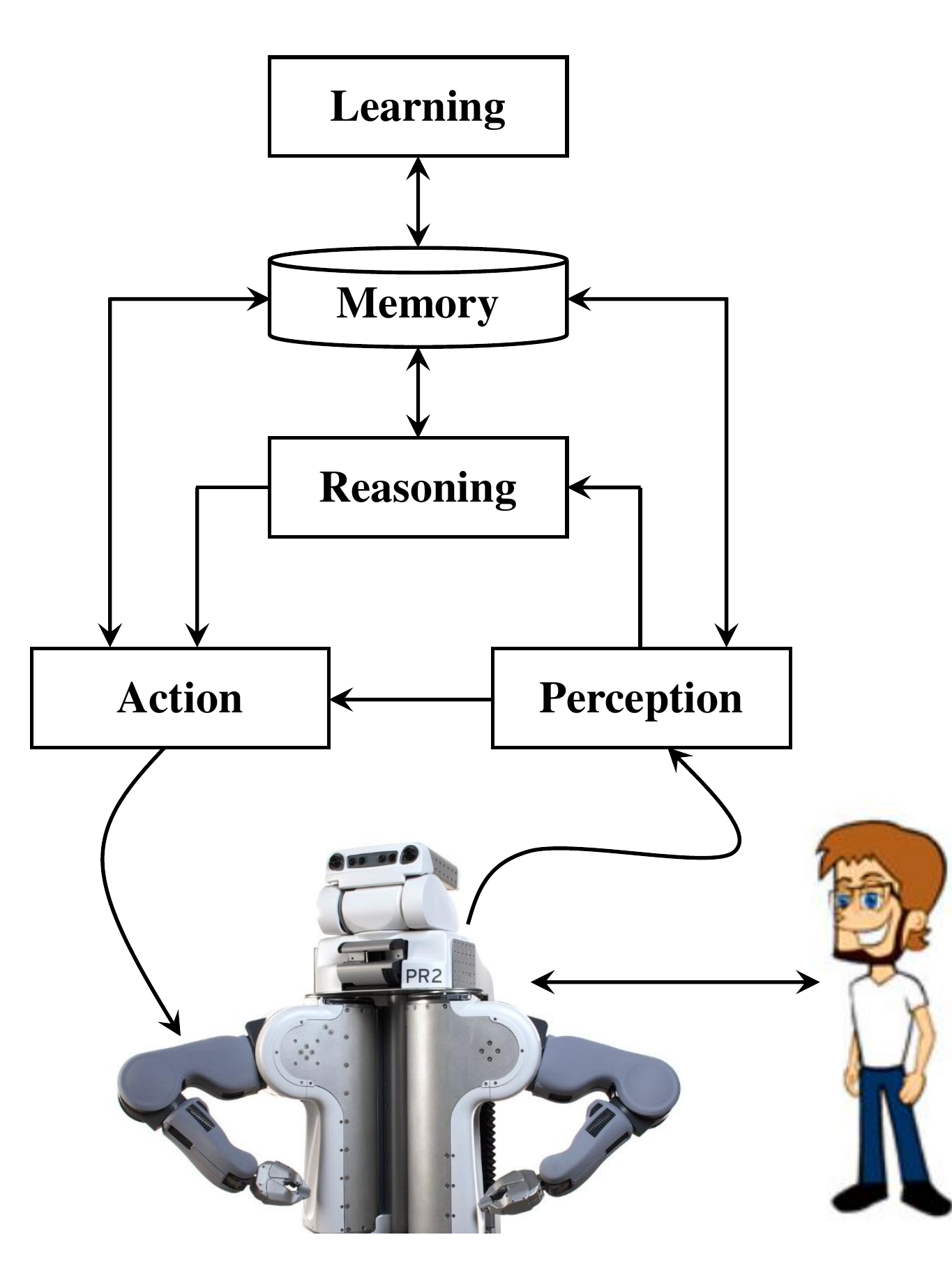}
    \vspace{-6mm}
  \caption{Abstract architecture for hybrid reactive-deliberative robots with a single memory system{, adapted from}~\cite{oliveira2014perceptual}.}
  \label{architecture}
  \vspace{-5mm}
\end{wrapfigure}
The abstract system architecture is depicted in Fig.~\ref{architecture}.
In this architecture, a \textit{Perception} component processes all momentary information coming from sensors, including sensors that capture the actions and utterances of the user. A \textit{Reasoning} component updates the world model and determines plans to achieve goals. An \textit{Action} component reactively dispatches and monitors the execution of actions, taking into account the current plans and goals. Finally, a Learning component, which typically runs in the background, analyzes the trace of foreground activities recorded in a \textit{Memory} component and extracts and conceptualizes possibly interesting experiences. The resulting conceptualizations are stored back in memory. It is worthwhile to mention that each component in such an abstract architecture is usually decomposed into a set of software modules.

This review of current state-of-art service robots will take a closer look at each piece in the pipeline of a service robot performing higher order tasks. Each core component will be reviewed based on recent works, describing the current state-of-the-art and identifying possible areas where improvements can still be made.
The remainder of this paper is organized as follows.  In Section~\ref{assistive_and_service_robots}, we review a set of state-of-the-art assistive and service robots. Afterward, in sections~\ref{perception}, \ref{grasping} and \ref{manipulation} the state-of-the-art research on {respecively perception, grasping and manipulation} is described with both positive developments and unsolved issues. Finally, in section~\ref{conclusion}, conclusions are presented and future works are discussed.

\section {Assistive and Service Robots}
\label{assistive_and_service_robots}

While a lot of developments have been made recently in each of the mentioned core components and in the field of service robotics as a whole, robotic servants do not yet live among us, helping us in our daily tasks. We believe that the underlying reason is that robots are usually painstakingly coded and trained extensively in advance to perform object perception and manipulation tasks in the right way. Therefore, the knowledge of such robots is fixed after the training phase, and any changes in the environment require complicated, time-consuming, and expensive robot reprogramming by expert users. {Additionally, the high costs of many of the more advanced robots, prices of tens of thousands or even higher being common (look for e.g. price estimates of PAL Robotics' TIAGo~\cite{pages2016tiago} or Fetch Robotics' Fetch~\cite{wise2016fetch}, robots which we will discuss below), further inhibits their widespread domestic use.} Although an exhaustive survey of assistive robotics is beyond the scope of this paper, representative works will be reviewed in this section.

Over the past decade, several projects have been conducted to develop robots to assist people in daily tasks. Most state-of-the-art service robots use classical object category learning and recognition approaches (i.e. offline training and online testing are two separate phases), where open-ended object category learning is generally ignored. Therefore, they work well for specific tasks, where there are limited and predictable sets of objects, and fail at any other assignment. In other words, the perceptual knowledge of these robots is static and they are unable to adapt to dynamic environments. Examples of such service robots that have demonstrated perception and action coupling include ARMEN~\cite{Armen}, El-E~\cite{El-E}, Busboy~\cite{Busboy}, TUM Rosie~\cite{Rosie}, 
TORO~\cite{TORO}, WALK-MAN~\cite{Walk-man}, ARMAR~\cite{ARMAR}, DLR’s Rollin'Justin~\cite{DLR}, { Automower~\cite{husqvarna}, Spot~\cite{bostondynamicsSpot}, Fetch~\cite{wise2016fetch} and TIAGo~\cite{pages2016tiago}}. {The characteristics of these robots are summarized in Table~\ref{table:list_of_reviewed_robots}}.

\begin{table}[!t]
    \caption{{Summary of reviewed service robots and their characteristics.}}
    \label{table:list_of_reviewed_robots}
    \resizebox{\linewidth}{!}{
    \begin{tabular}{|c|l|}
        \hline 
        \rowcolor{gray!30} \textbf{Robots} & \textbf{Characteristics}  \\\hline\hline
        El-E~\cite{El-E} & - 3D vision based on a stereocamera, a monochrome camera and a laser scanner \\& - One arm manipulator \\& -  Two fingered gripper with force sensors \\&- Wheels that allow for mobility \\&- Can be directed by a human using a laster pointer \\\hline
        ARMEN~\cite{Armen}  & - 2D vision based on a greyscale camera \\&- One arm manipulator \\&- Three fingered gripper \\&- Wheels that allow for mobility \\&- Visual memory building \\&- Interaction with humans via a virtual conversational agent\\\hline
        Busboy~\cite{Busboy}  & - Consists of two cooperating robots \\&- 2D vision based on external camera \\&-  One arm manipulator on one of the robots \\&- Three fingered grasper \\&- Both are mobile wheeled robots \\\hline
        TUM Rosie and James~\cite{Rosie}  & - Consists of two cooperating robots \\&- 3D vision based on a combination of a tilting laser and a ToF camera \\&- Both robots have two manipulators \\&- One has two fingered grippers while the other has 4 fingered human-like grippers \\&- Both are mobile wheeled robots \\&-  Able to learn new tasks from instructions on the internet\\\hline
        TORO~\cite{TORO}  & - 3D vision based on a stereo camera and depth sensor \\&- Two manipulators \\&- Two anthropomorphic hands \\&- Bipedal motion  \\\hline
        WALK-MAN~\cite{Walk-man}  & - 3D vision based on a stereo camera and a laser sensor \\&- Two manipulators \\&- Two anthropomorphic hands \\&- Bipedal motion  \\\hline
        ARMAR~\cite{ARMAR}  & - 3D vision based on a stereo camera, a laser sensor and a RGB-D sensor \\&- Two manipulators \\& - Two anthropomorphic hands \\&- Wheels that allow for mobility \\&- Human robot interaction via pose estimation, verbal and physical interactions \\\hline
        DLR’s Rollin'Justin~\cite{DLR}  & - 3D vision photo mixer device cameras \\&- Two manipulators \\&- Two anthropomorphic hands \\&- Wheels that allow for mobility   \\\hline
        Fetch~\cite{wise2016fetch}   & - 3D vision based on RGB-D cameras \\&- One arm manipulator \\&- Two fingered gripper \\&- Wheels that allow for mobility   \\\hline 
        TIAGo~\cite{pages2016tiago}  & - 3D vision based on RGB-D cameras \\&- One arm manipulator (two arms are also possible) \\&- Anthropomorphic hands \\&- Wheels that allow for mobility \\&- Several modular options \\\hline
        Boston Dynaqmics' Spot~\cite{bostondynamicsSpot}  & - 3D vision based on stereo camers \\&- Quadrupedal motion \\&- Interaction with humans via a tablet application \\&- Several modular options (e.g. a manipulator with a two fingered gripper)\\\hline
        HERB\cite{HERB}  & - 2D/3D vision based on two RGB cameras  \\&- On arm manipulator \\&- Three fingered gripper \\&- Wheels that allow for mobility \\&- Interaction with humans via voice recognition and a laser pointer   \\\hline
        Roomba~\cite{iRobot}   & - 3D mapping using low resolution camera \\&- Wheels that allow for mobility \\&- Interaction with humans via mobile application \\\hline
        Braava~\cite{iRobot} & - 3D mapping using low resolution camera \\&- Wheels that allow for mobility \\&- Interaction with humans via mobile application\\\hline 
        Husqvarna's Automow~\cite{husqvarna} & - Perception based on electronic and radio signals, as well as soft collisions \\&- Wheels that allow for mobility \\&- Interaction with humans via mobile application \\\hline

    \end{tabular}
    }
\end{table}

In the ARMEN project, Leroux et al.~\cite{Armen} proposed a mobile assistive robotics approach providing advanced functions to help care for elderly or disabled people at home. This project mainly involves object manipulation, knowledge representation, and object recognition. The authors also developed an interface to facilitate the communication between the user and the robot. Jain et al.~\cite{El-E} presented an assistive mobile manipulator, named EL-E, that can autonomously pick objects from a flat surface and deliver them to the user. The user should provide the location of the object to be grasped by the robot by pointing at the object with a laser pointer.

In another work, a busboy assistive robot has been developed by Srinivasa et al.~\cite{Busboy}. In particular, they propose a multi-robot assistive system, consisting of a Segway mobile robot with a tray and a stationary Barrett WAM robotic arm. The Segway robot navigates through the environment and collects empty mugs from people. Then, it delivers the mugs to a predefined position near the Barrett arm. Afterwards, the arm detects and manipulates the mugs from the tray and loads them into a dishwasher rack. The vision system of busboy is designed for detecting a single object type (mugs). Furthermore, because there is a single object type (i. e. mug), they computed the set of grasp points off-line.

The work on two robots cooperating to solve a pancake making task \cite{Rosie} shows the feasibility of having one or multiple robots perform a task they were not specifically programmed for. In this experiment, it is shown that by giving a robot information on objects and low-level tasks, used in conjunction with a task planning module, these robots can make their own order of operations from a set of instructions downloaded from the internet. The instructions are vague and directed towards humans with some prior knowledge of the objects and operations involved. When an instruction mentions preheating the pancake maker, for example, this could mean turning it on, or it could also require plugging in the power cord. A robot making its own order of operations will need to take all these differences into account. This opens up routes for further dynamic task resolution, giving a robot the {capacity} to solve a wide variety of problems, in contrast with most robotic projects, which work to solve a small range of tasks which are specifically defined for that purpose.

Another interesting robot which has been in development for a while is TORO \cite{TORO}, a continuation of the DLR's Rollin' Justin \cite{DLR}. This robot was developed to explore the possibilities of torque-controlled joints in humanoid robots. While several forms of torque control are specified, the electrical drive units with torque measurements are argued to be most effective for service robots in a household setting. The torque feedback from the limbs while moving gives the robot an additional layer of safety when interacting with humans as any movement can be adjusted not just by input from the cameras, but also from the limbs themselves.

Combining both actuators and torque to balance passive and active adaptation to the environment{,} WALK-MAN \cite{Walk-man} was developed. This robot was designed in response to a call for disaster relief robots and, to perform well in these hectic environments, focuses on being able to move effectively in very rough terrain. In this work, it is argued that, while quadrupedal or wheeled robots are often more stable, they are incapable of properly traversing significantly imbalanced terrain. To further diminish the effect of uneven terrain, the combination of active and passive adaptation was developed. This resulted in a very robust bipedal robot.

While multiple of these robots have been developed for work in a human environment, ARMAR \cite{ARMAR} is specifically developed to collaboration with humans. The focus of this robot is on applicability in a human work place and working side by side with humans. To achieve this, the reach and carrying capacity have been developed significantly. ARMAR has a reach of {1.3 m} and can carry {10 kg} in one arm, even at maximum reach. Most importantly, ARMAR has been developed with the ability to detect human behaviour in an attempt to recognize when a human is in need of help. This is done through pose-tracking of the human, which can be combined with speech commands, giving the robot information on when and how to help. It can also estimate the task a human is doing and determine whether it is a task normally done by multiple people. If so, the robot can step in as the second person.

{All of the service robots discussed so far were relatively small-scale research projects, however, service robots on a wider commercial scale have also been developed. The most well known and widespread of such robots are probably small wheeled household robots, such as Roomba~\cite{iRobot}, Braava~\cite{iRobot}, which are respectively an automated vacuum cleaner and mop developed by iRobot~\cite{iRobot}. Additionally, similar robots such as robotic lawnmowers, like Husqvarna's Automow~\cite{husqvarna}, are also in relatively widespread use. While among the more expensive household products, they are competitively priced among other high end items and have {seen use} in many households around the world. However, they are all relatively simple, only have to operate in one specific type of area, are low to the ground and slow moving (which greatly simplifies navigation and obstacle avoidance) and can only perform one specific task.}

Nevertheless, more advanced and versatile commercial service robots have also been developed. Boston Dynamics' Spot~\cite{bostondynamicsSpot} is an example of a highly mobile quadrupedal robot. With its on-board 360 3D cameras it can traverse a large range of terrains, like cluttered household environments or steep staircases. Additionally, it can recover from crashes and collisions and can operate in a large variety of conditions. With it's range of add-ons, such as a robotic arm and more powerful sensory modules, it can be applied to a diverse range of tasks, such as the fetching of items, surveying, or remote monitoring of patients. Some of its main drawbacks are that it is relatively short and can thus not reach all household objects{,} and it has a limited battery life of only 90 minutes.

{While being less mobile, taller, armed robots such as Fetch Robotics' Fetch~\cite{wise2016fetch} and the more customizable PAL Robotics' TIAGo~\cite{pages2016tiago}, do not have these same problems. They can have battery lives of 8-10 hours, allowing them to operate as long as a human would on a typical working day. These robots are both very similar, having shared features like a laser scanner at the base, a 3D RGB camera on the top and an extendable torso, to reach even higher places with their arm (or possibly two arms in the case of TIAGo). They can be applied to a large range of service tasks, as well as research projects. However, they, as well as Spot, still suffer from similar (computational) limitations as many of the aforementioned robots, not yet being able to function independently in many dynamic environments. Additionally, while they are affordable for companies, healthcare institution, and research groups, they are still outside the budget of most households.}

\begin{figure*}[b]
\centering
\includegraphics[width=\linewidth]{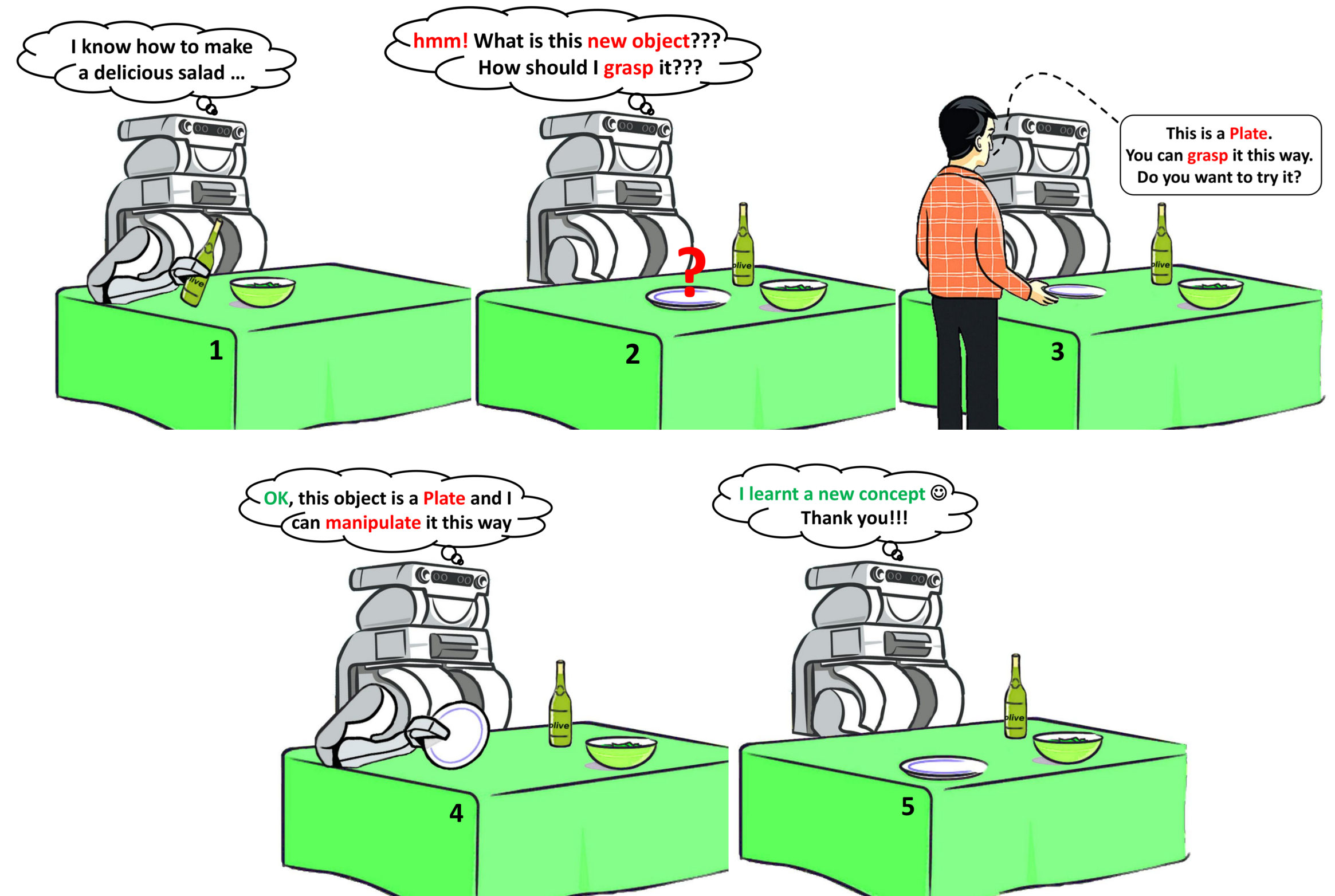}
\caption{ An illustrative example of a ``\textit{make\_a\_salad}'' task with a PR2 robot: In this scenario, the robot faces a new object while making a salad. A user then teaches the ``\textit{Plate}'' category to the robot and demonstrates a feasible grasp for the object. This way, the robot conceptualizes new concepts and adapts its perceptual motor skills over time to different tasks. }
\label{make_a_salad}
\end{figure*}

In order for robots to be useful day to day, they need to understand and make sense of the spaces where we live and work, and adapt to new environments by gaining more and more experiences. In such dynamic domains, robots are expected to increasingly interact and collaborate closely with humans. This requires new forms of machine intelligence. Towards this goal, \textit{open-ended learning} must be properly understood and addressed to better integrate robots into our society. In human cognition, learning is closely related to memory. Wood et al.~\cite{wood2012review} presented a thorough review and discussion on memory systems in animals as well as artificial agents, having in mind further developments in artificial intelligence and cognitive science. Cognitive science also revealed that humans learn to recognize object categories and grasp affordances ceaselessly over time \cite{harnad2017cognize,lebedev2002insights}. This ability allows adapting to new environments by enhancing knowledge from the accumulation and conceptualization of new experiences. Inspired by this theory, service robots should approach 3D object recognition and manipulation from a long-term perspective and with emphasis on domain open-endedness. Moreover, we have to incorporate intermittent robot-teacher interaction, which is an outstanding feature in human learning~\cite{illeris2018comprehensive}. This way, non-expert users will be able to correct unexpected actions of the robot and quickly guide the robot toward target behaviors. For example, consider the robotic ``\textit{make\_a\_salad}'' task as depicted in Fig.~\ref{make_a_salad}. In this example, the robot faces a new object while making a salad. The robot then asks a user to teach the category of the object and demonstrate how to grasp it. Such situations provide opportunities to collect training instances from online experiences, and the robot can incrementally update its knowledge rather than retraining from scratch when a new task is introduced or a new category is added.

To achieve this, several cognitive robotics groups have started to explore how to learn incrementally from past experiences and human interactions to achieve adaptability. Besides, since service robots receive a continuous stream of data, several methods have been introduced that use open-ended learning for object perception.
In \cite{skovcaj2016integrated}, a system with similar goals is described. Faulhammer et al., \cite{faulhammer2016autonomous}, presented a perception system that allows a mobile robot to autonomously detect, model, and re-recognize objects in everyday environments. They only considered isolated object scenarios. Of course, actual human living environments can be very cluttered, that is why Srinivasa et al. \cite{HERB} introduced HERB. HERB is an autonomous mobile manipulator that can navigate through a dynamic and highly cluttered environment. Furthermore, it is able to search for and recognize objects in this clutter and is able to manipulate a large range of objects, including constrained objects like doors. The main drawback of HERB is that its error recovery is completely hand-coded, which does not interact sufficiently with other modules (e.g.\ planning modules). As a result, HERB can become stuck in cycles in which after recovery of an error, it keeps performing the behaviours that lead to the same error over and over again.  

In the RACE project \cite{RACE,oliveira20163d}, a PR2 robot demonstrated effective capabilities in a restaurant scenario including the ability to serve coffee, set a table for a meal, and clear a table. The aim of RACE was to develop a cognitive system, embodied by a service robot, which enabled the robot to build a high-level understanding of the world by storing and exploiting appropriate memories of its experiences. 
The X company also launched a similar project, recently, called the everyday robot project\footnote{\href{https://x.company/projects/everyday-robots}{https://x.company/projects/everyday-robots}}. The main goal of this project is to develop a general-purpose learning robot that can safely operate in human environments, where things change every day, people show up unexpectedly, and obstacles appear out of nowhere.

{Another interesting thing to address is that it might well be possible that multiple service robots operate in one shared environment (think for example about smaller robots such as the vacuum clearer and/or mob robots mentioned before working in a household with one or even multiple of the larger more versatile robots also discussed above, such as in \cite{Rosie}). Cloud services allow for a centralized way to control or guide all of these robots while also reducing the need for expensive computational resources on the robots themselves. The DAvinCi framework~\cite{davinci} is an example of this, in which robots communicate with a centralized server through the internet (or other forms of wireless communication), which also controls the sharing of information between the different robots. Another, more general approach is provided by the Internet of Robotic Things~\cite{IoT}, which allows for vaster and more intricate networks of control which can also directly couple to and exchange information with for example disembodied sensors, mobile phones, personal computers and even any number of humans. Such cloud services offer many benefits, most notable of which is perhaps that it allows for highly parallel computation. However, it also has several downsides, such as control lag issues (especially prevalent in households with a slow internet connection) and due to it's connective nature, a strong vulnerability for hacking compared to self controlled robots. For a more detailed overview of ins and outs and the pros and cons of the use of cloud services in robotics see e.g. the review \cite{cloudReview1}, or the more recent\cite{cloudReview2}.}

\begin{figure}[!b]
\includegraphics[width=\linewidth]{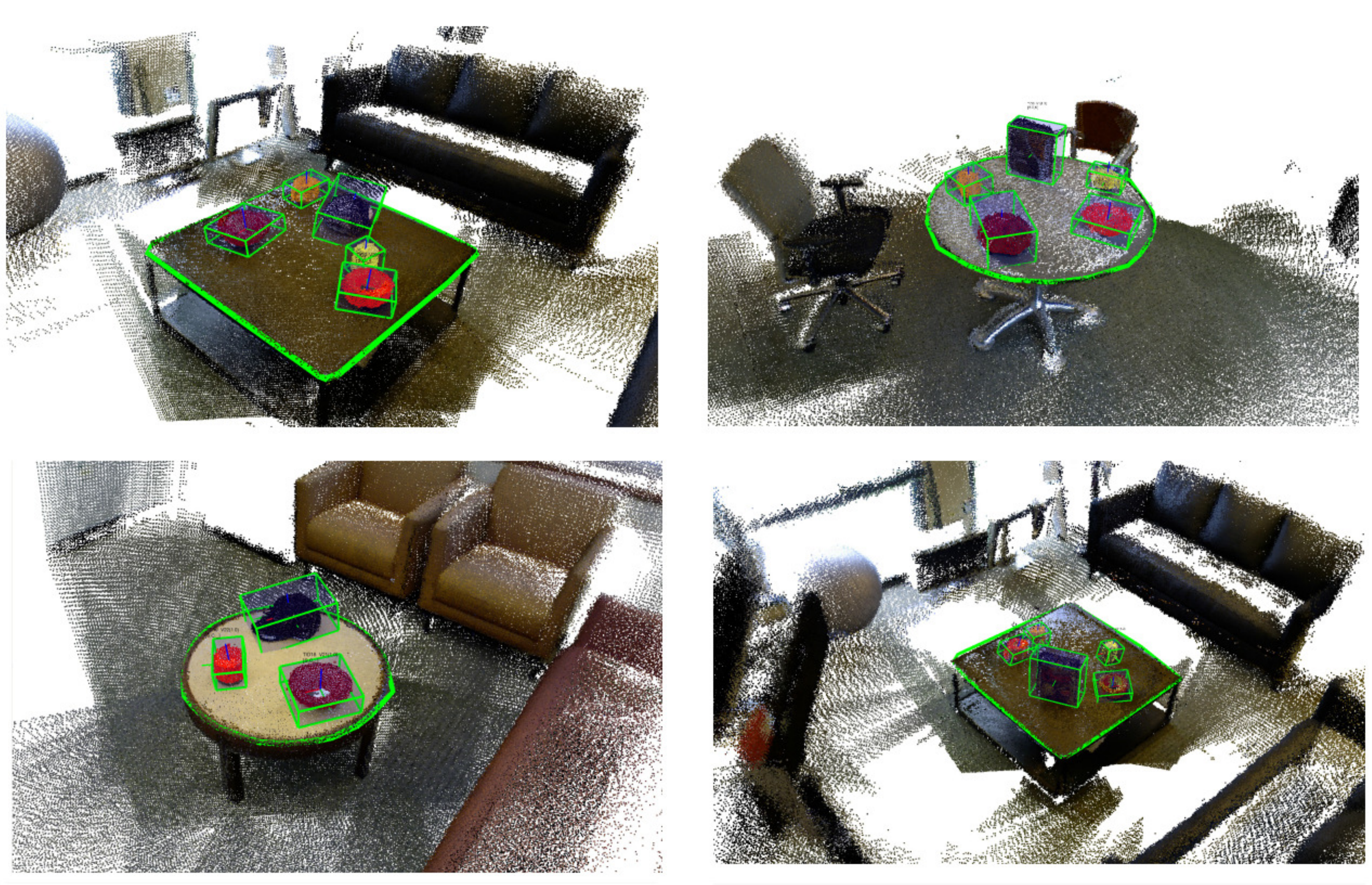}
\vspace{-4mm}
\caption{Examples of object segmentation in isolated objects scenarios: detected object candidates are shown by different bounding boxes and colors.}
\label{isolated_objects}
\vspace{-2mm}
\end{figure}

\section {Object Perception and Perceptual Learning}
\label{perception} 
{Following the recent release of inexpensive 3D sensing devices, which simultaneously record RGB and depth information, object perception became one of the most researched subjects in robotics and computer vision communities. Object perception comprises a wide range of tasks, including object detection, object category learning, and recognition. Over the last decade, researchers have focused more on deep learning-based approaches that have been mainly facilitated by the vast increase in available computational power.} The main goal of object perception is to detect the pose and recognize the label of any objects the robot needs to interact with in order to help human users. Furthermore, the output of object perception is required for motion planning not only to specify the pose of a target object for the manipulation purposes but also to localize all objects that can be considered as obstacles for the current task. 
Additionally, it is important that a robot knows about its work-space and accessible regions within the work-space. In the following subsections, we discuss the recent advances in object detection, open-ended object category learning and recognition, and the alternate forms of perceptions for service robots.

\subsection{Object detection}
Object detection and pose estimation are crucial for robotics applications and recently attracted attention of the research community~\cite{sock2017multi}. Many researchers participated in public challenges such as the Amazon picking challenge\footnote{\href{https://www.amazonrobotics.com}{https://www.amazonrobotics.com}} to solve multiple object detection and pose estimation in a realistic scenario. There are two different mechanisms that are often employed for real-time object detection purposes.

\begin{figure}[!b]
\includegraphics[width=\linewidth]{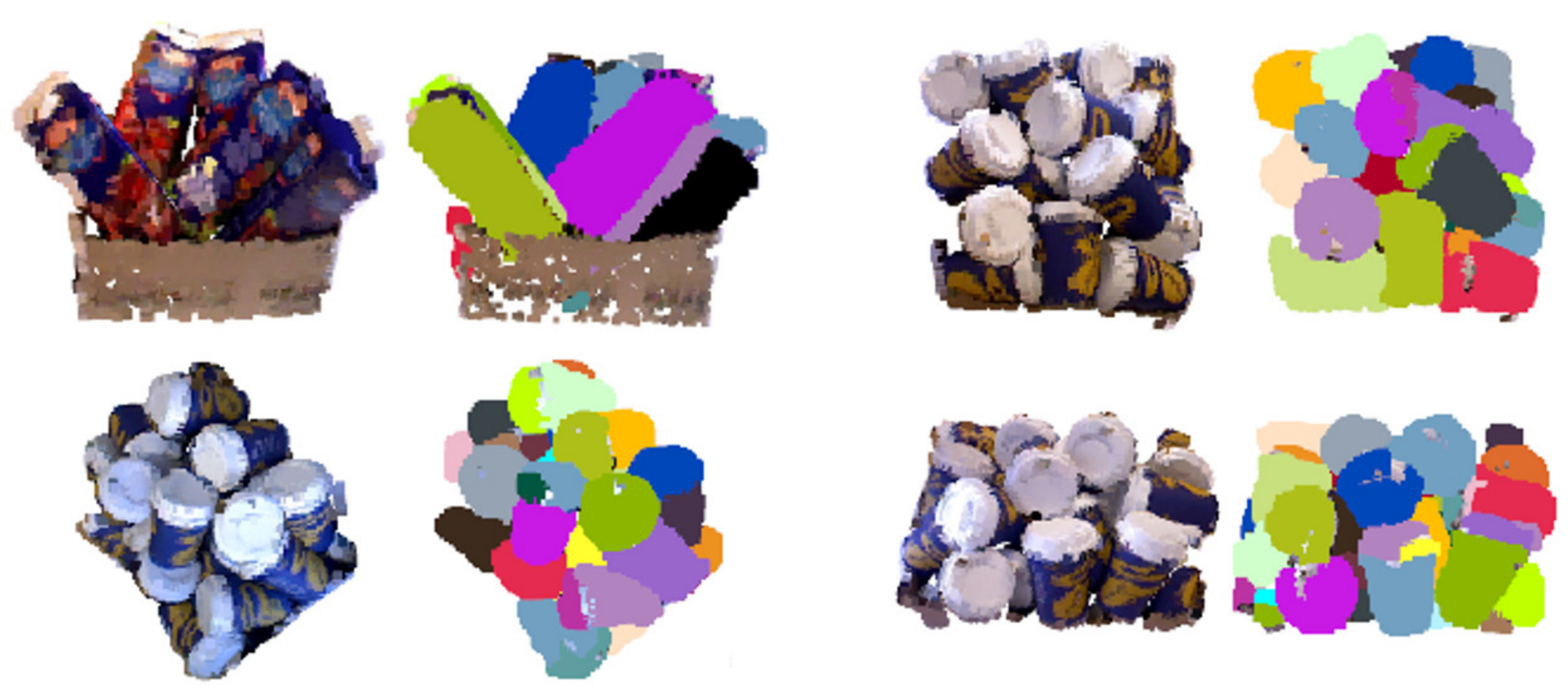}
\vspace{-5mm}
\caption{Four examples of object segmentation in pile scenarios: detected objects are shown by different colors.}
\vspace{-4mm}
\label{pile_of_objects}
\end{figure}

\begin{figure}[!t]
\includegraphics[width=\linewidth]{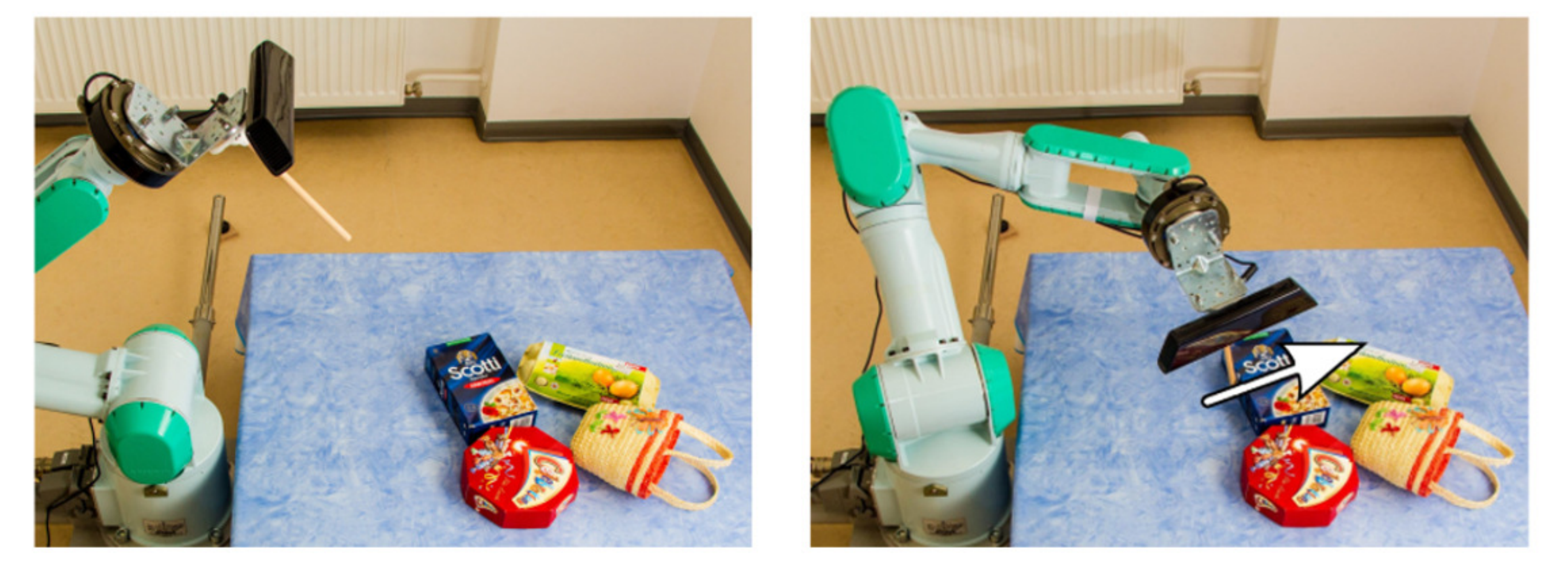}
\vspace{-4mm}
\caption{Examples of active object segmentation in pile of objects scenarios: (\textit{left}) The robot
observes the scene, obtaining a point cloud {from} one perspective. (\textit{right}) The robot decides to push the object from the bottom to segment them. The robot repeats these step to singulate all objects (adapted from \cite{van2014probabilistic}).}
\label{active_segmentation}
\vspace{-2mm}
\end{figure}

In the first group of approaches, an object detection approach is paired with an approach for object classification. These approaches initially detect objects and place a bounding box around them. Each detected object is then classified{~\cite{szegedy2013deep,redmon2016you,bochkovskiy2020yolov4}}. Four examples of isolated table-top objects are shown in Fig.~\ref{isolated_objects}. In general, separating object detection and object recognition is a suitable strategy for service robots since it allows robots to detect never-seen-before objects \cite{redmon2016you,zhao2019object,kasaei2018towards,GOOD,kasaei2019orthographicnet,kasaei2015interactive}. A drawback, however, is that only detected objects will be classified. In a household environment, a robot may frequently encounter a pile of objects such as a clutter of toys in the living room, tidying up a messy dinning table, or multiple unused objects stacked in a box in the garage. This complicates the object detection process, because some objects can be occluded by and overlapping with other objects. There are two sets of approaches that can handle \textit{pile segmentation}. 
The first set of approaches mainly use clustering algorithms to segment objects using the curvature of the surface normals and/or colour information \cite{colourSegmentationGPoint,2DColour,regionGrowing2D} (see Fig.~\ref{pile_of_objects}). The other set is based on active segmentation. In other words, such approaches initially segment the scene to specify a set of object candidates and then, manipulate/push object candidates one by one to be isolated from the pile of objects~\cite{chang2012interactive}. For example, Van Hoof et al.~\cite{van2014probabilistic} presented a part-based probabilistic approach for interactive object segmentation. They tried to minimize human intervention in the sense that the robot learns from the effects of its actions, rather than human-given labels (see Fig.~\ref{active_segmentation}). In another work, Gupta and Sukhatme~\cite{gupta2012using} explored manipulation-aided perception and grasping in the context of sorting small objects on a tabletop. They presented a pipeline that combines perception and manipulation to accurately sort the bricks by color and size. 

\begin{figure}[!t]
\includegraphics[width=\linewidth]{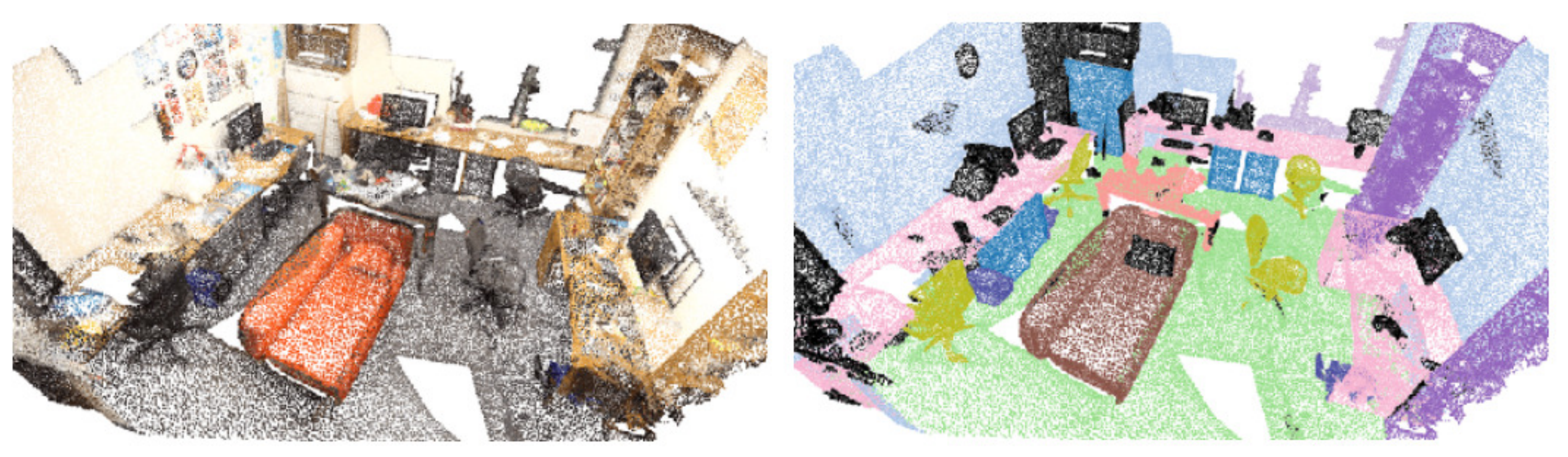}
\vspace{-4mm}
\caption{Examples of object (scene) segmentation: (\textit{left}) a
3D input pointcloud; (\textit{right}) a network prediction; The colors of
the points represent the object labels (adapted from \cite{choy20194d}).}
\label{scene_segmentation}
\vspace{-2mm}
\end{figure}

{In the second group of real-time object detection, we have \textit{semantic segmentation} approaches that classify each point of a scene to a particular class \cite{long2015fully} (i.e., end-to-end object detection, see Fig.~\ref{scene_segmentation}). This group does not have the mentioned problem of the first group \cite{choy20194d,kim2019integration,oliveira2018efficient,long2015fully,deepSegmentation,mo2019partnet}. However, in some real-world scenarios, semantic segmentation approaches are computationally expensive, and cannot produce consistent labeling for each point. In other words, a point may get no label or several labels depending on how many segmented regions contain that point. In these approaches, each segmented region corresponds to (\textit{i}) a salient part of an object, (\textit{ii}) an entire object, or (\textit{iii}) a group of objects belonging to the same category.} 

An example of a semantic segmentation approach on object parts segmentation ({case \textit{i}}) is given by SpiderCNN~\cite{xu2018spidercnn}. As the name suggests, it uses {Convolutional Neural Networks (CNNs)}, however, with one important adaption. Normal CNNs cannot straightforwardly be applied to point clouds data, since it is not stored in a regular grid, which is required for such CNNs. SpiderCNN instead uses a set of irregular parametrized convolutions, which can be applied to point clouds directly. A similar approach is used by PointCNN \cite{li2018pointcnn}, which instead uses x-transforms to be able to extend convolutions to point clouds. Alternative approaches that do not use CNNs have also been explored. Take for instance PointNet \cite{qi2017pointnet}, it initially processes all points independently and identically with a set of input and feature transforms. Eventually it combines point features by applying one layer of max pooling. While performance is on par with other state-of-the-art approaches, its drawback is that due to its layers of independent processing, it does not capture local features of the spatial distribution of points well. To this end PointNet++ \cite{qi2017pointnet++} was introduced. It borrows ideas from CNNs and applies the architecture of PointNet recursively on incrementally growing local regions of the input space. Unlike the original pointNet, this now allows it to capture local features on a variety of scales.

Several attempts have been recently made to not only segment objects by category, but also by different instances in a category~\cite{instanceSegmentation} (case \textit{ii} and \textit{iii}). Some approaches use top-down methods to handle instance segmentation problem based on detect-then-segment scheme (e.g., R-CNN~{\cite{RCNN}}, and its variants such as Mask RCNN~{\cite{maskRCNN}}, Fast RCNN~{\cite{fastRCNN}}, Faster RCNN~{\cite{fasterRCNN}}, PANet~{\cite{wang2019panet}}, Mask Score RCNN ~{\cite{maskScoreRCNN}}, etc.). These approaches first extract the bounding boxes of each object instance and then perform binary segmentation inside each bounding box to distinguish the object from the background. This group of approaches is quite slow to be used by a service robot in real-time.  There has been a wave of studies on single-stage instance segmentation; notable works are SOLOv2~{\cite{SOLOv2}}, SOLOv1~{\cite{SOLOv1}}, CenterNet~{\cite{centernet}}, and FCOS~{\cite{FCOS}}. It {is} worth mentioning that these methods are usually faster and more accurate than detect-then-segment based approaches~{\cite{SOLOv2}}. Furthermore,  all target objects are segmented, which could provide valuable information for the motion/task planning purposes.

While end-to-end deep learning based object detection can achieve high accuracy, it generally requires a very large number of training examples{,} and training with limited data usually leads to poor performance \cite{redmon2016you,zhao2019object}. Catastrophic forgetting is another important limitation of deep learning approaches \cite{kirkpatrick2017overcoming,kemker2018measuring}. Furthermore, these approaches are unable to learn new categories in an online fashion. {For further discussion on major drawbacks and challenges of using deep neural networks in lifelong learning scenarios, we refer the reader to the recent review paper on continual lifelong learning with neural networks by Parisi et al.,~\cite{parisi2019continual}.} It is worth mentioning that for path planning purposes, the mentioned limitations are not that important, since the robot mainly needs to avoid objects, furniture or humans, which can be divided in{to} relatively few categories. However, to be of help, a service robot must be able to manipulate all sort of small household items accurately, which come in a seemingly endless number of categories. 

Another problem with many current 3D object detection approaches is the inability to detect transparent and highly reflective objects. Since glass and shiny metal objects are commonplace in every household, this is an issue that surely needs to be addressed. S. Sajjan et al.,~\cite{ClearGrasp} took a step towards addressing such limitations for object grasping and manipulation. In particular, they proposed a deep learning approach, named ClearGrasp~\cite{ClearGrasp}, for estimating accurate 3D geometry of transparent objects from a single RGB-D image {(which is a combination of an RBG image and a corresponding depth image)} for robotic manipulation.

\subsection{Object category learning and recognition}

A typical household environment contains objects belonging to a large number of categories. In such domains, it is not feasible to assume one can anticipate and preprogram everything for robots. Therefore, autonomous robots must have the ability to execute learning and recognition concurrently. Several methods have been introduced that allow for open-ended learning of new categories. In such approaches, the introduction of new categories can significantly interfere with the existing category descriptions. To cope with this issue, memory management mechanisms, including salience and forgetting, can be considered~\cite{kirkpatrick2017overcoming}. The main approaches can be divided into two different categories, according to what type of object representation they use. On the one hand, there are approaches that do still use deep learning based techniques~\cite{hariharan2017low,ullrich2017selecting,kasaei2019orthographicnet}. They use networks that have been pretrained on a large dataset, and, therefore, are already able to extract a lot of useful features from images. Additional training and recognition on new objects and categories is approached in a number of different ways, some use few-shot learning~\cite{hariharan2017low,gidaris2018dynamic,oreshkin2018tadam}, others use one-class support vector machines~\cite{krawczyk2015one} and random forests~\cite{ristin2014incremental}, and simple instance-based learning is also combined with nearest neighbor classification \cite{hariharan2017low,ullrich2017selecting,kasaei2019orthographicnet}. Other learning approaches have also been considered, such as the use of autoencoder-based representation learning~\cite{tschannen2018recent,zhao20193d}, Bag of Words~\cite{bagOfWords,kasaei2018towards,kasaei2015adaptive}, and topic modeling \cite{Local-LDA,kasaei2016hierarchical,kasaei2016concurrent}. {To accommodate the significant computational power some of these methods need, some have proposed the use of cloud services to replace costly on-board hardware~\cite{cloudPerception1,cloudPerception2}. Note that a benefit of this is that such cloud services are compatible with any of the discussed methods (including the ones that will follow below) and it could even use several of these methods in parallel.}

On the other hand, there are approaches that instead use hand-crafted features for object recognition \cite{GOOD,VFH,PFH,FPFH,ESF}. They use either global or local features of the objects that can be extracted in a variety of ways. The representation of an object is usually given by a histogram of features. Concerning category formation, an instance-based learning (IBL) approach is usually adopted, in which a category is represented by a set of views of instances of the category. When the robot is presented with a new object, it compares the representation of the new object to the representation of all instances of all categories by, for example, a simple nearest neighbor classifier. Therefore, each instance-based object learning and recognition approach can be seen as a combination of a particular object representation, similarity measure~\cite{Distance} and classification rule~\cite{KasaeiColorConstancy2020}. One advantage that instance-based learning has over other methods of machine learning is its ability to adapt the model to previously unseen data. The disadvantage of this approach is that the computational complexity can grow with the number of training instances. The computational complexity of classifying a single new instance is $O(n)$, where $n$ is number of instances stored in memory. Therefore, these systems must resort to experience management methodologies to discard some instances {and} prevent the accumulation of an impractically large set of experiences. Salience and forgetting mechanisms can be used to bound the memory usage. These mechanisms are also useful for reducing the risk of overfitting to noise in the training set. Another advantage of the instance-based approach is that it facilitates incremental learning in an open-ended fashion.

There are some works that adapt model-based learning (MBL) that are often contrasted with IBL approaches. In particular, IBL considers category learning as a process of learning about the instances of the category while MBL is a process of learning a parametric/non-parametric (Bayesian) model from a set of instances of a category. In other words, each category is represented by a single model. The IBL approaches propose that a new object is compared to all previously seen instances, while the MBL approaches propose that a target object is compared to the model of categories. Therefore, in the case of recognition response, MBL approaches are faster than IBL approaches. In contrast, IBL approaches can recognize objects using small number of experiences, while MBL approaches need more experiences to achieve a good classification result. Therefore, training is very fast in IBL approaches, but they require more time in the recognition phase. Another disadvantage of IBL approaches is that they need a large amount of memory to store the instances. In MBL, new experiences are used to update category models and then the experiences can be forgotten immediately. The category model encodes the information collected so far. Therefore, this approach consumes a much smaller amount of memory when compared to any IBL approach.

\subsection{Fine-grained object recognition}
In human-centric environments, fine-grained (very similar) object categorization is as important as basic-level categorization. A problem of the above approaches is that categories that are very similar might be hard to distinguish. Such categories could for example be different items of cutlery, different dog, cat or other pet breeds, a variety of box shaped objects (like food container boxes, tissues, a stack of paper, etc) or different writing utensils. Attempts have been made to tackle this issue by introducing fine grained object recognition \cite{kasaei2019look,gao2013learning,zhang2016weakly}. Fine-grained object recognition takes into consideration small visual details of the categories that are important to distinguish them from similar categories.  In the case of food boxes, think for example about the print on the boxes. Fine-grained and basic-level recognition are important in different domains and, when combined, a trade-off is made between the computational benefits of basic-level recognition and the accuracy of fine-grained recognition. Most existing approaches for fine-grained categorization heavily rely on accurate object parts/features annotations during the training phase. Such requirements prevent the wide usage of these methods. However, some works only use class labels and do not need exhaustive annotations. Geo et al. \cite{gao2013learning} proposed a Bag of Words (BoW) approach for fine-grained image categorization by encoding objects using generic and specific representations. This approach is impractical for an open-ended domain (i.e., large number of categories) since the size of object representation is linearly dependent on the number of known categories. Zhang et al.~\cite{zhang2016weakly} proposed a novel fine-grained image categorization system. They only used class labels during the training phase. This work completely discarded co-occurrence (structural) information of objects, which may lead to non-discriminative object representations and, as a consequence, to poor object recognition performance. Several types of research have been performed to assess the added-value of structural information. Kasaei et al.~\cite{Local-LDA} proposed an open-ended object category learning approach just by learning specific topics per category. In another work~\cite{kasaei2019look}, an approach is proposed to learn a set of general topics for basic-level categorization, and a category-specific dictionary for fine-grained categorization.

\begin{figure}[!b]
\centering
\includegraphics[width=0.9\linewidth]{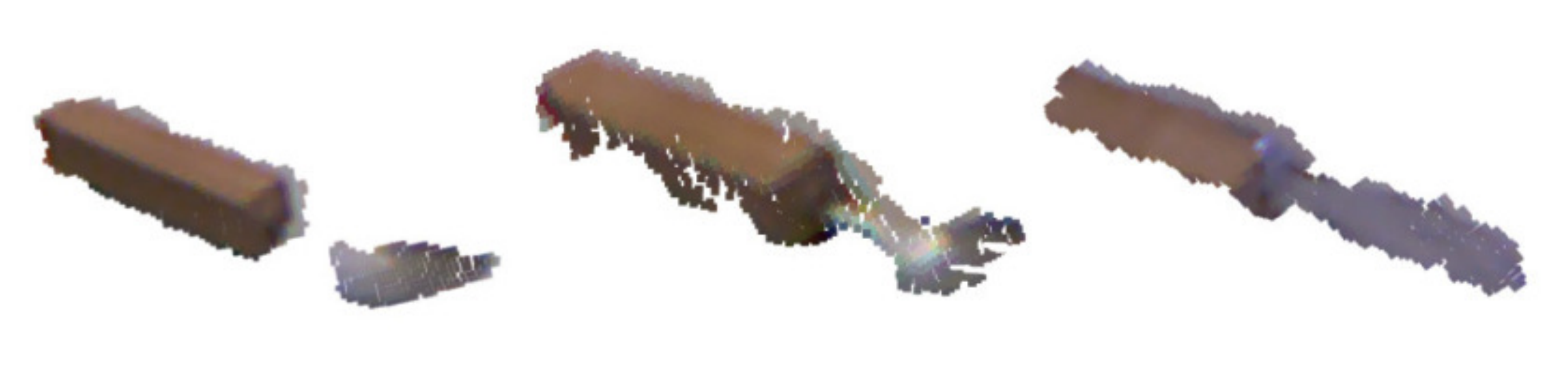}
\vspace{-4mm}
\caption{A set of very similar (fine-grained) cutlery objects: (\textit{left}) an object view of a spoon, (\textit{center}) an instance of a fork, and (\textit{right}) an object view of a knife object.}
\label{fine_grained_objects}
\vspace{-4mm}
\end{figure}
Object detection and recognition performance can be improved by considering the context in which objects appear \cite{luddecke2019context,contextMicrosoft,contextBasedRecognition,mottaghi2014role}.
In a house, certain categories of related items are often placed together. This information can be used to improve the processing of related items. For example, chairs and tables often appear together, so the presence of one could be used as a cue to expect the presence of the other as well. Additionally, information can be used to distinguish objects from similar categories that appear in different environments. A pen and a screwdriver, for example, have a fairly similar shape, however, a pen is likely to appear on or near a desk in a home office or a table in the living room, while a screwdriver is expected to be around other tools such as hammers and wrenches that are more likely to be found in a garage or near a fuse box.

In active perception scenarios, whenever the robot fails to recognize an object from the current view point, the robot will estimate the next view position and capture a new scene from that position to improve the knowledge of the environment. This will reduce the object detection and pose estimation uncertainty. Towards this end, Mauro et al.~\cite{mauro2014unified} proposed a unified framework for content-aware {Next-Best-View} selection based on several quality features such as density, uncertainty, and 2D and 3D saliency. Using these features, they computed a view importance factor for a given scene. Kasaei et al.~\cite{kasaei2018perceiving} proposed a novel Next-Best-View prediction algorithm to improve object detection and manipulation performance. First a given scene is segmented into object hypotheses and then the {Next-Best-View} is predicted based on the properties of those object hypotheses. In another work, Biasotti et al.~\cite{li2013sketch} approached the problem of defining the representative views for a single 3D object based on visual complexity. They proposed a new method for measuring the viewpoint complexity based on entropy. Their approach revealed that it is possible to retrieve and to cluster similar viewpoints. Doumanoglou et al.~\cite{doumanoglou2016recovering} used class entropy of samples stored in the leaf nodes of a Hough forest to estimate the Next-Best-View. Some researchers have recently adopted deep learning algorithms for {Next-Best-View} prediction in active object perception. For instance, Wu et al.~\cite{wu20153d} proposed a deep network namely 3D ShapeNets to represent a geometric shape as a probabilistic distribution of binary variables on a 3D voxel grid.

\subsection{Alternate forms of perceptions}
Cognitive scientists showed that humans' vision is not an independent process and it is closely coupled to other forms of perception \cite{stein1993merging,ernst2004merging,eckert2008cross}. It could therefore be desirable to also explore different forms of perception in robotics. Furthermore, it could be interesting to investigate a way to couple these forms of perception, similar to what we see in humans. This could be especially useful in the case where fragmented information in the different modes of perception could be combined to form a clearer picture.

Take, for example, the case of robotic perception via touch. A robot cannot only use vision, but also touch to identify objects \cite{luo2017robotic}. This could help with identifying transparent or reflective objects, which was previously identified as a problem area and is something many forms of visual object detection currently cannot {handle}. It could also be very useful for a robot to be able to detect typical household sounds, which has received surprisingly little attention so far \cite{kertesz2018common}. Additionally, auditory and visual perception could be improved by interchange of information. Research in this respect has largely focused on sound localization \cite{2.5dVisualSound,gao2019co,gao2018learning,ephrat2018looking,zhao2018sound,zhao2019sound}. This entails that, given a video input with a corresponding audio, a system tries to identify which sounds in the audio channel correspond to which objects in the video scene. Take for example the MONO2BINAURAL \cite{2.5dVisualSound} DNN. Given a single-channel audio signal together with an accompanying source video, it uses spatial information in the video to convert the audio signal into two channels, each of which represents the signal received in one ear. This improved double audio channel then also provides spatial information of the sounds, which is clearly related to sound localization. It is therefore not surprising that the representation learned by this network was successfully extended to the task of sound localization. The {CO-SEPARATION} \cite{gao2019co} architecture is very similar, {however}, it separates the audio into different sources instead of two spatial channels. By doing this it also learns to match the sound of the same object type appearing in different videos.

While sound localization can already be useful in itself for robots to, for example, look at the person who is talking to it, to the {authors'} best knowledge no attempts have been made to leverage this separated and localised audio to aid in the recognition of objects or events. For instance, imagine the case of a whistling kettle, or something more serious like the thump of a human falling and hitting the ground, possibly injuring themselves. It is important for the robot to be able to identify where the sound comes from. Furthermore, it is just as important that the robot should be able to recognise the sounds, and the objects that produce them, and take action if necessary. 

{Additionally, aside from touch, hearing and vision, humans and many animals also have a sense of smell and taste. It could be useful to also impart these abilities on robots. Especially robotic olfaction has received an increasing amount of attention over the years \cite{smellReview,behaviourBasedSmeller,rightDirectionSmell,nextBestSmell}. Research here is mostly focused on detecting harmful chemicals and localising their sources. An often used example, that is common place in society, is the detection and tracking as of a gas leak. In a household environment such leaks could result in dangerous situations. It would therefore be good if household robots could find such leaks and alert the proper authorities, or possible even fix the leaks in certain scenarios (think of e.g. an unlit stove top being left on, which could be turned off by a robot with a grasper with relative ease). Additionally, this could also be beneficial in other situations, such as the case of domestic spills of toxic cleaning products, that could form a risk especially to young children.} 

{Another route of research has focused on using artificial olfaction as a means to monitor and improve food quality~\cite{grapeNose,NoseIndustry}, which is often combined with taste sensors as well~\cite{foodQuality,EQuality}, taste sensors in themselves also having been applied in the same manner~\cite{grapeTongue,tasteFabrication}. The sensors involved with this research are often called E-noses and E-tongues and they have mainly been applied outside of the field of robotics. Nevertheless, the potential uses for this technology in robotics are vast. Robotic cooks could use taste and smell to improve their cooking, to check if a food product has spoiled or is still edible, to, along the same vein, check if any harmful substances are present in the food or even to check the ripeness of fruit. Now efforts are already being made to apply these kinds of technology in the field or robotics. Take for example~\cite{fingerTaste}, in which a robotic hand is equipped with electrochemical sensors on three of its fingers. One finger being able to measure spiciness, another sweetness and the remaining one sourness. However, the level of taste achieved by such applications is still very limited.}

{Another downside of current taste and smell sensors is that they usually saturated quickly and then take some time before they are ready to be used again, although efforts are being made to reduce such issues~\cite{bioSmeller}. It's currently also difficult to compare different methods, both because they are applied in a variety of different scenarios and because no widespread performance standards exist~\cite{smellTasteStandards}. Fortunately this field of research is still young and further development of this technology seems inevitable. In the long term, ideally smell and taste would be combined with all other forms of perception discussed above, to form one unified framework, so that each form of perception can benefit from information of the others (note that outside of robotics some efforts in this direction are already being made~\cite{tasteAndVision,TasteAndVision2}). }

\section{{Learning Affordance and Object Grasping}}
\label{grasping}
{As we discussed in the introduction section,} \textit{Object grasping and manipulation} stand at the core of a successful service robot, without these abilities the robot cannot provide many of the necessary services. There are a number of problems inherent to this task. For example, the robot needs to accurately detect the pose of objects, recognise the class of objects, and find the best location to grasp them in real-time. Besides, any planned movements need to avoid self collision and collision with the environment. Furthermore, the more complex a robotic arm and gripper are, the more complicated these problems become, taking more computation time to find adequate solutions. Both accuracy and computation time are the two major factors to balance when writing algorithms for these purposes. {A great deal of work has been done to improve the quality of object grasping and reduce the execution of motions in object manipulation using machine-learning techniques. There are still several areas and specific subjects that have not yet received a lot of attention. In the following subsections, we will point out these gaps in the state-of-the-art since further advances in robots' ability to grasp and manipulate objects in different situations could have enormous societal impact, from picking packages in logistic centers to assistive robots helping disabled people to gain independence in their daily activities, e.g., cleaning the table after eating a meal.}

\subsection{Grasp point detection}

Detecting stable grasp points for previously unseen objects in real-time is one of the main challenges for object grasping. Towards finding proper grasping points of objects, many different techniques for both teaching and learning processes can be used. Trial and error is not often utilized due to the large risk of damaging the robot, but it is able to be employed to learn multiple good grasps \cite{ficuciello2019hand,miller2004graspit}. Some approaches use object segmentation, but they differ significantly from the approaches mentioned in the previous section{. These} often have the disadvantage of being relatively inflexible and not generalize well to objects outside of the set it has been trained on. Matching known grasps in a manner of instance-based learning is possible as well \cite{kopicki2016one}. Many recent techniques share the same pipeline to find a set of stable grasp points for the given object. First the grasping candidates are sampled from the point cloud (or image) of the object, then the candidates are ranked through, for example, a CNN~\cite{mahler2017dex}. The best candidate then gets grasped in either an open-loop or a closed-loop fashion. Grasp point detection approaches can be broadly classified into two categories, \textit{analytical} approaches and \textit{empirical} approaches. While analytical approaches rely on kinematic and dynamic formulations to choose proper end-effector configurations (i.e., {the} position and orientation of hand and fingers), empirical approaches use data-driven learning algorithms to transfer grasps from 3D model databases to a target object. Empirical methods can be further classified based on whether the grasp configuration is being computed for \textit{known}, \textit{familiar} or \textit{unknown} objects. The underlying reason for this classification is that prior knowledge about objects determines how grasp candidates are generated and ranked. For more details on data-driven grasp synthesis, we refer the reader to the surveys of Bohg et al.~\cite{bohg2013data} and Sahbani~\cite{sahbani2012overview}.

Deep learning methods have recently achieved the largest advancements in grasping {previously unseen objects. }
Although an exhaustive survey is beyond the scope of deep learning methods for grasp point detection, we review some representative works here.  Most of these approaches however use an adaption of the CNN architectures designed for object recognition~\cite{johns2016deep}. Additionally, they often sample and rank potential grasps individually~\cite{lenz2015deep}. Together, this can cause exceedingly long computation times which makes them unsuitable for real-time closed-loop grasps. Mahler et al.~\cite{mahler2016dex} proposed the first Dexterity Network (Dex-Net). Since then they have proposed three version of Dex-Net architectures. Together with the DexNet architectures they also proposed the growing Dex-Net dataset. The Dex-Net 1.0~\cite{mahler2016dex} uses a CNN with {an} Alex-Net architecture~\cite{krizhevsky2012imagenet} for predicting the labels of the given object first and{,} then, the probability of forced closure under uncertainty in object pose, gripper pose and friction coefficient to obtain the quality of a grasp. {Note use is made of cloud services both to store the large dataset and to perform computations on up to 1,500 virtual cores simultaneously to drastically decrease computation times (as much as 3 orders of magnitude).} The Dex-Net 2.0~\cite{mahler2017dex} uses a Grasp Quality Convolutional Neural Network (GQ-CNN) to evaluate grasp candidates for a certain object. In this approach, first, a discrete set of antipodal candidate grasps is sampled from the image space. Then, these grasp candidates are forwarded to the GQ-CNN to find out the best grasp point. The architecture of Dex-Net 3.0~\cite{mahlerdex3} does not differ from the architecture from the Dex-Net 2.0, and the main difference lies in the use of a suction cup in the Dex-Net 3.0, whereas the Dex-Net 2.0 uses parallel-jaw gripper. The Dex-Net 4.0~\cite{mahler2019learning} was created for a robot with a suction cup and a parallel-jaw gripper. The advantages of suction cups are that they can reach into narrow spaces and pick up an object with a single point of contact. In difference to earlier versions, the Dex-Net 4.0 not only trains on grasping objects in separation, but it is also trained to grasp objects in pile scenarios. 

\begin{figure}[!t]
\includegraphics[width=\linewidth]{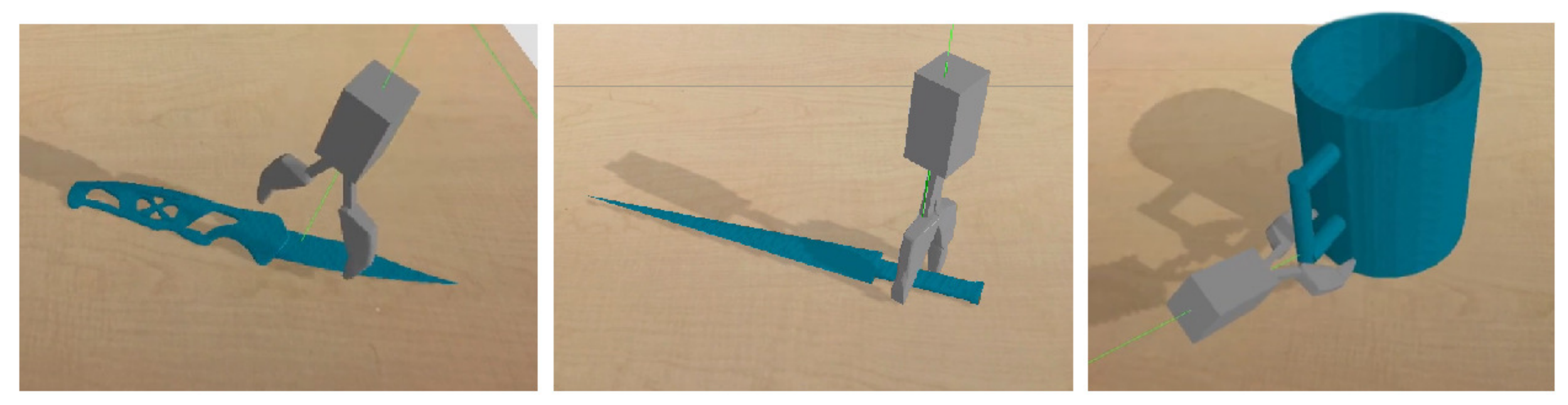}
\vspace{-3mm}
\caption{Results of grasp point detection on three different objects: (\textit{left}) Without considering object affordance, the robot detects a graspable point on the blade region of the knife; (\textit{center} and \textit{right}) By considering objects' affordances, the robot detects a suitable grasp point for both objects.}
\vspace{-2mm}
\label{fig:grasping}
\end{figure}

Generative Grasping Convolutional Neural Network (GG-CNN)~\cite{morrison2018closing,morrison2019learning} seeks to improve on the mentioned drawbacks. GG-CNN is an object-independent grasp synthesis method which can be used for closed-loop grasping. GG-CNN predicts the quality and pose of grasps at every pixel through a one-to-one mapping from the depth image. This is in contrast with most deep-learning grasping methods which use discrete sampling of grasping candidates and have long computation times. Furthermore, such online analysis of objects allows for more precise and potentially faster grasps. These methods vastly reduce the amount of possible grasps, by highlighting the locations where the grasps are of the highest quality. In another work~\cite{zeng2018robotic}, {a method of} online grasp generation has been addressed. Both~\cite{zeng2018robotic} and GG-CNN approaches use neural networks to generate maps of grasp quality. For the former, grasp points for the two end-effectors at various angles are generated. For the latter, maps for grasp quality, angles and width are generated. 

A robot could benefit from affordance detection of parts of a single object to reduce the complexity in finding the locations of high quality grasps~\cite{myers2015affordance,do2018affordancenet,myers2015affordance,qian2020grasp}.
Certain objects have a clear place to be grabbed, for instance the handle of a knife, the ear of a mug or the handle of a pan. These can be identified in this manner. On the contrary, there are also objects that should not be grasped in a certain area. Considering again the case of a knife, its blade could damage the robot's end-effector when grasped. Softer skin-like end-effectors \cite{elango2015review}, which are important for delicate manipulation of objects, are especially at risk (see Fig.~\ref{fig:grasping}). Care should also be taken with plates, cups, bowls and similar objects. They may contain food or liquids. Manipulating filled containers is already a difficult task \cite{moriello2018manipulating}, a bad grasp can make the manipulation more difficult than it has to be. Naturally, touching food or liquids with the end-effector should be avoided as well. 

An affordance for a certain grasp is generally synonymous with a high grasping quality, be it for a specific end-effector, angle or other purpose. In \cite{antanas2019semantic} and \cite{kaiser2018autonomous}, objects' point clouds are semantically segmented by a {rule-based} system. Based on what the robot is tasked with, {a} specific segment of the object may be more suitable to be grasped than others. As a bonus, the sizes and shapes of segments are features, allowing the system to classify objects. This combination of grasp affordance and knowledge of semantics, creates a system that is robust in handling objects for a variety of tasks. Besides, affordance prediction can provide some additional information about other objects \cite{sun2010learning,hermans2011affordance,kaiser2018autonomous}. For example, it can tell the robot on which furniture other objects can be placed, or determine if an object can be picked up or not, etc. It is worthwhile to mention that affordances do not make any claims about the category of objects in a scene, but rather try to predict the functionality of objects either by visual identification \cite{hermans2011affordance} {or by sorting them into hierarchical structures~\cite{kaiser2018autonomous}. The latter can optionally be cobbled together using belief maps sourced from multiple different perceptual inputs. On the opposite end of the spectrum, we may see the desire to get as much information from a single point of view as possible, in which case ~\cite{qian2020grasp} provides a neural network based solution similar to Dex-Net for findign the ideal grasp with proper affordances.} Instance-based learning with affordances \cite{zeng2018robotic} and applying affordances to segments of objects \cite{antanas2019semantic} are other possible options.

\subsection{Object grasping} 
{Object grasping remains a challenging task because of the variety of knowledge from different domains required to pull it off. Using inverse kinematics (IK) it is possible to eliminate impossible or undesirable configurations of the manipulator such as singularities and collisions.}
 A good grasp has a high quality, a measure of how stable a grasp is, and has to be somewhere on the visible part of the object. A large body of recent efforts has focused on solving 4-DoF ($x$, $y$, $z$, $yaw$) grasping, where the gripper is forced to approach objects from above \cite{zeng2018robotic,morrison2019learning,mahler2017dex,GGCNN}.

{A major drawback of these approaches is that it restricts the range of possible interactions with objects.} For instance, they are not able to grasp a horizontally placed plate. Worse still, the knowledge of robots is fixed after the training stage, in the sense that the representation of the known grasp templates does not change anymore. A household situation is prone to change, cabinets may be stocked differently and implements may be lost and gained. To avoid impossible tasks, the degrees of freedom of an end-effector cannot be so restricted. These drawbacks call for more advanced and seamless approaches for object grasping.

Recently some research groups have taken a step towards addressing these issues. Qin et al.,~\cite{qin2019s4g} studied this problem in a challenging setting, assuming that a set of household objects from unknown categories are casually scattered on a table. They proposed a learning framework to directly regress 6-DoF grasps from the entire scene point cloud in one pass. In particular, they compute a per-point scoring and pose regression method for 6-DoF grasp.
In another paper, a single view was used for grasp generation by Kopicki et al.,~\cite{kopicki2019learning}. This works similarly {to} the direct regression of point clouds. The main obstacle with such a method is that the back side of objects is generally unknown and represent missing information. Challenging situations, those where the back side needs to be grasped, are thus harder to correctly resolve. By combining generative models and using new ways to evaluate contact points, a higher rate of success is achieved in these challenging situations.
In another work, Murali et al.,~\cite{grasp6D} proposed an approach to plan 6-DoF grasps for any desired object in a cluttered scene from partial point cloud observations. They mainly used an instance segmentation method to detect the target object. To generate a set of grasp points for the object, they follow a cascaded approach by first reasoning about grasps at an object level and then checking the cluttered environment for collisions. 
{To add to the previous work, Haustein et al.~\cite{haustein2017integrating} address the issue of grasping in an environment where obstacles are obscuring the most direct path to the desired object. The proposed solution involves creating a tree structure of possible approach strategies and then rapidly exploring the tree to find the optimal pose.}
Yet another example of integrating both arm {motion} planning and grasp planning is {shown} in a paper by Fontanals et al.~\cite{fontanals2014integrated}. This work instead focuses on combining a method to grasp an object with a grasp pose that minimizes the chance of collision with the rest of the environment as well. In the context of a slightly more dexterous robot (SpaceJustin), the points of contact on an object are considered independently when building a grasp.

These methods are partial solutions to the general problem of moving from 4 to 6 degrees of freedom. A larger search space for grasps means that there are more grasps that are in conflict with the environment. A larger search space also means more computation time is required to find good grasps and eliminate bad ones. By focusing on regression of point clouds and single views, Qin et al. and Kopicki et al. avoid a large computational pitfall. Likewise, the method of Murali et al. is able to avoid scene collisions by clever segmentation and analysis. 

For the methods discussed, the end-effectors are still rather simple. While fast grasping with a simple gripper is relatively easy, doing so with more complicated and also more dextrous grippers is again harder and more computationally intensive. Another option is a vacuum end-effector which functions like a suction cup. Like grippers, these have similar requirements for a successful grasp. Both require a high quality grasp, but while grippers require a stable pinching area, suction cups require a stable suction area. This difference in suitable grasp location for vacuum end-effectors allows them to manipulate objects in ways that grippers cannot.

\begin{figure}[!b]
\vspace{-4mm}
\includegraphics[width=\linewidth]{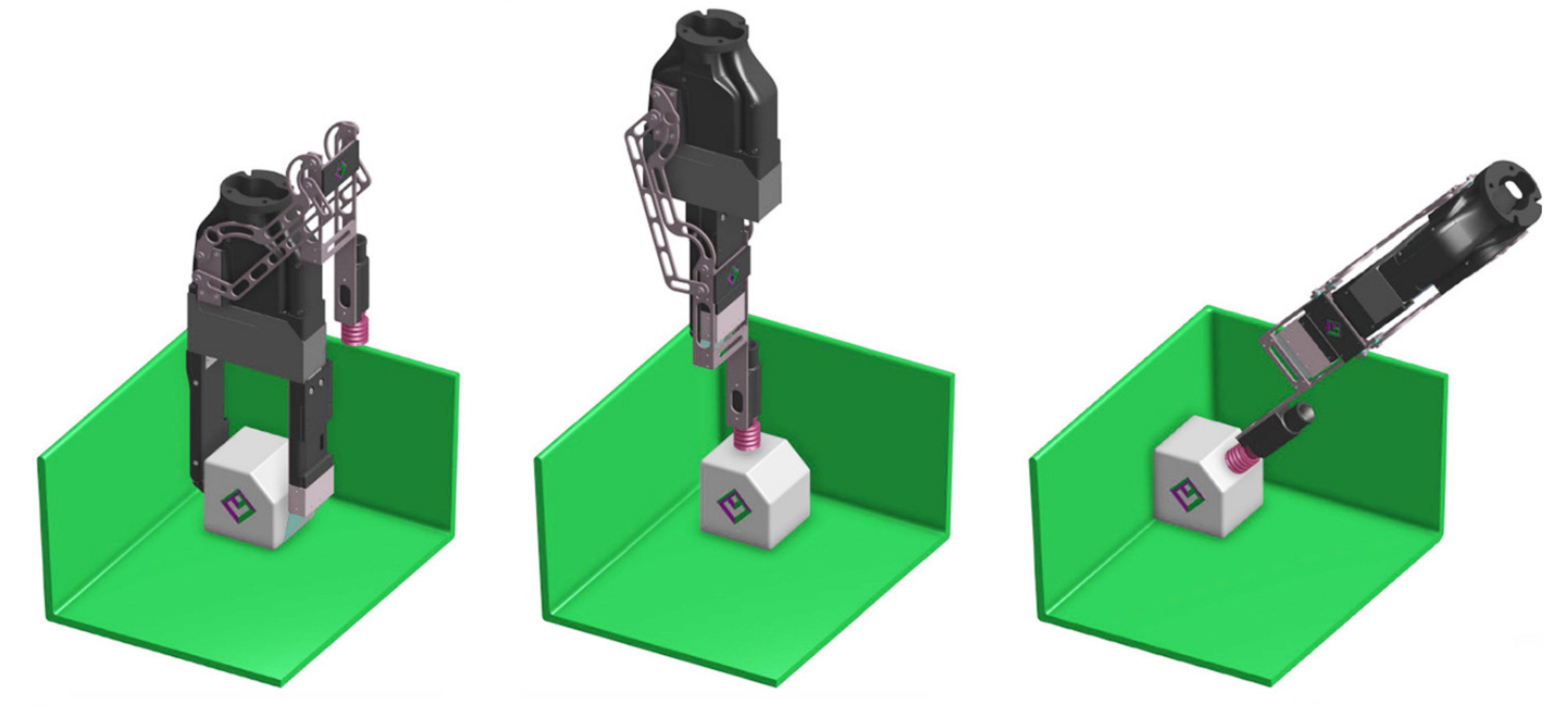}
\vspace{-2mm}
\caption{An example of multi-functional gripper that can be used for grasping objects in different situations: (\textit{left}) In this situation, the object is graspable vertically using the two-finger
parallel-jaw gripper since there is enough space between walls and the object. This type of grasping is mainly used to pick up objects with
small, irregular surfaces such as baskets, table-top objects and tools;
(\textit{center}) Some objects can be robustly grasped using a suction cup  gripper as shown in this figure. This type of grasping is robust for objects with large and flat surfaces, e.g., books and boxes. (\textit{right}) In some situations, it is necessary to grasp an object from an specific direction mainly due to environmental constraints or application requirement. For instances, this type of grasp is used to grasp objects resting against walls, which may not have suction-able areas from the top~\cite{funkhouser2019robotic}.}
\label{multi_functional_gripper}
\end{figure}

As we saw before, Mahler et al.,~\cite{mahlerdex3} created Dex-Net 3.0, a large dataset for 3D object grasping specifically for vacuum-based end-effectors. This dataset is mainly used to train a grasp quality classifier. Therefore, simple objects and typical everyday objects are grasped with a high rate of success while the most complicated of objects are able to be successfully grasped over half of the time.
While a vacuum end-effector is highly suitable for work in a factory, a delicate gripper is still preferred for the manipulation of the softest of objects, as well as those objects that lack any proper suction points. The opposite is also true, objects that are more easily picked up with a suction cup{,} or that are not able to be picked up with a gripper{,} may warrant the inclusion of a vacuum end-effector. To ensure the proper completion of any task, a service robot may have to be equipped with both. 
Instead of using two robotic arms with two different types of gripper, Zeng et al.,~\cite{funkhouser2019robotic} developed an interesting multi-functional gripper, consisting of a suction cup and a parallel-jaw gripper, to allow robots to robustly grasp objects from a cluttered scene without relying on their object identities or poses. Such mechanisms enable a robot to quickly switch between suction cup and parallel-jaw gripper to grasp different type of objects in various positions (see Fig.~\ref{multi_functional_gripper}).

There are also {various} anthropomorphic robotic hands. The shadow dexterous hand \cite{shadowrobotcompany} is an almost fully actuated robotic hand that is modeled very closely after human hands. There even exists an upgraded version which boasts touch and vibration sensors on the fingertips of the device. For a more integrated approach, iCub~\cite{metta2010icub} is a fully anthropomorphic robot that was designed for human-like interactions with its environment. Compared to the shadow dexterous hand, the {hands of the }iCub robot are softer and less actuated. {They are} covered in a soft skin-like material to aid in the delicate handling of objects. On the extreme end of the spectrum, there is the RBO Hand-II~\cite{deimel2016novel}. While still anthropomorphic, it's individual fingers are more {akin} to tentacles than human fingers. It is an extremely soft and relatively simple hand, lacking any joints. Nevertheless it is still able to grasp a variety of objects.

{This seems to suggest that the simplicity of a gripper does not necessarily inhibit it from being successful at various tasks.} This is also shown in a review of various robotic hands \cite{piazza2019century}, where there has been a rise in simpler but effective grippers. While complex end-effectors can be used to do complex tasks, a simpler gripper may be able to perform it just as effectively. This leads back to the vacuum end-effector which is far more effective in certain circumstances than any anthropomorphic robotic hand.

While soft or under-actuated robotic hands are not required for delicate operation, it generally depends on the task at hand. Industrial robots are often fully actuated and rigid because, while the actions themselves are often not complicated and in a highly structured environment, high precision is still required. For service robotics, the tasks are more varied{, yet some of these tasks do require} similar precision {(e.g. a task like opening a locked door, window or cupboard, requiring precise manipulation of a key)}. For such an environment, the end-effector may specify suitable tasks instead of the other way around. Having multiple end-effectors at its disposal then increases the amount of tasks that the robot is suitable for. To this end, a combination of different grippers such as in Zeng et al.,~\cite{funkhouser2019robotic} would be a proper approach to create a capable service robot.

\subsection{Open-ended grasp learning}

\begin{figure}[!t]
\includegraphics[width=\linewidth]{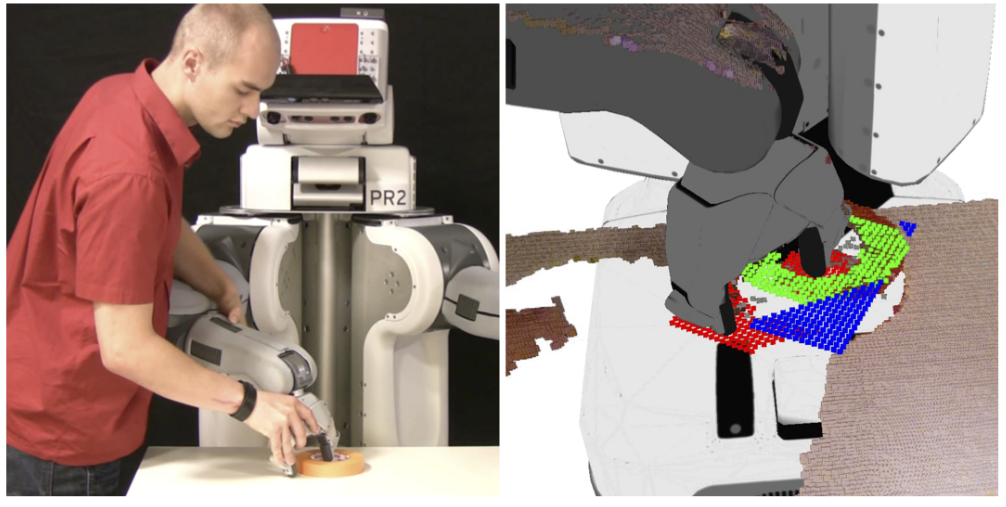}
\vspace{-4mm}
\caption{An example of kinesthetic teaching: (\textit{left}) a user demonstrates a feasible grasp to the PR2 robot; (\textit{right}) Extracted template heightmap and gripper-pose are used to train the proper grasp position for the given object (adapted from ~\cite{herzog2014learning}).}
\label{kinestetic_teaching}
\end{figure}

While progress has been made towards open-ended learning, it remains a big problem that needs to be addressed further. Current approaches are only able to successfully learn a limited percentage of all objects that can be found in a household. For service robots which need to be able to help those in need, significant chance of failure to recognise and grasp objects, for important objects like medicine containers, could be very problematic. Towards addressing this issue, some researchers used kinesthetic teaching to teach a new grasp configuration, including the position and orientation of the arm relative to the object and the finger positions~\cite{kasaei2019interactive,herzog2014learning,shafii2016learning}. As shown in Fig.~\ref{kinestetic_teaching}, an instructor teaches an appropriate end-effector position and orientation using the robot's compliant model. After performing the kinesthetic teaching, visual features of the taught grasp region (e.g., heightmaps~\cite{herzog2014learning}) are extracted and stored as a grasp template in the grasp memory. 

In another work, Kasaei et al.~\cite{kasaei2019interactive}, formulated grasp pose detection as a supervised learning problem to grasp familiar objects. Their primary assumption was that new objects that are geometrically similar to known ones could be grasped in a similar way. They used kinesthetic teaching to teach feasible grasp configurations for a set of objects. The target grasp point is described by (\textit{i}) a spin-image~\cite{johnson1999using}, which is a local shape descriptor, and (\textit{ii}) a global feature, which represents the distance between the grasp point and the centroid of the object. To detect a grasp configuration for a new object, they initially estimate the similarity of the target object with all known objects, and then, try to find the best grasp configuration for the target object based on the grasp templates of the most similar object.

Most of these examples have dealt with only grasping. To be truly considered dexterous, robotic systems should have the capacity to do complicated tasks. A small step in the right direction is the ability to prevent objects from slipping and falling \cite{huang2017intelligent}. The adjustment of already held objects, such as by using the individual fingers \cite{sundaralingam2019relaxed}, is a required stepping stone to advanced object handling. Moving from one robotic arm to two, synchronized or not \cite{mirrazavi2018unified}, truly opens up the way to manipulate equipment made for human use. A fair part of the equipment that service robots have to work with, is made for humans. This means that advanced object manipulation, as well as the other advances, are required for the development of capable service robots.

One such example is to use the manipulators available to the robot in potentially not the most intuitive way for a robot, by not considering a grasping or attaching action to be the end goal of the manipulation. Some approaches that address this are described by Eppner et al.~\cite{eppner2015planning} and a somewhat more focused look is provided by Hang et al.~\cite{hang2019pre}. Starting with Eppner et al. we can see that objects may have complex functions when considered as a whole but we can also apply much simple motions to them using moves that are not quite grasps, so pushing and sliding motions. These allow us to manipulate an object without actually grasping. An example of this would be bracing an unstable object against a hard surface in order to get a stable grasp point on it. In the paper this is described as using environmental constraints as a basis for manipulation. The paper by Hang et al. takes this a step further, or rather focuses on a particular sub-section of this problem addressing a point brought up earlier. Picking up objects too flat to grab by sliding them into a position where a grasp is possible, such as the edge of a table. There is however, a drawback to  all of this. If the manipulator used by the robot is not able to provide any feedback as to the obstacles it comes up against, it runs the risk of damaging components, so such a strategy for manipulation may even be undesirable in some situations. 

{Addressing the second part of the concerns raised earlier, related to handling objects made for humans; post-grasp actions, i.e. how to re-position an object in a hand with either a the object holding gripper itself and/or a separate arm.} For this, several methods have been proposed at present. First, there is a proposed method of adapting to an object which may be slipping from the grip. This is addressed in two studies by Huang et al.~\cite{huang2017intelligent} and Hang et al.~\cite{hang2016hierarchical} where the first proposes varying the pressure placed on the object and the latter rather focuses on adding a step for re-positioning fingers on a dexterous gripper in order to attain a better grasp on the object. A separate, more recent study shows the use of reinforcement learning to re-position or re-grip an object in a single dexterous gripper~\cite{andrychowicz2020dextrous}. This is done with the example of re-orienting a cubical object through a combination of finger position features and visual feedback.


\section{Object Manipulation}
\label{manipulation}

\begin{figure}[!t]
\includegraphics[width=\linewidth]{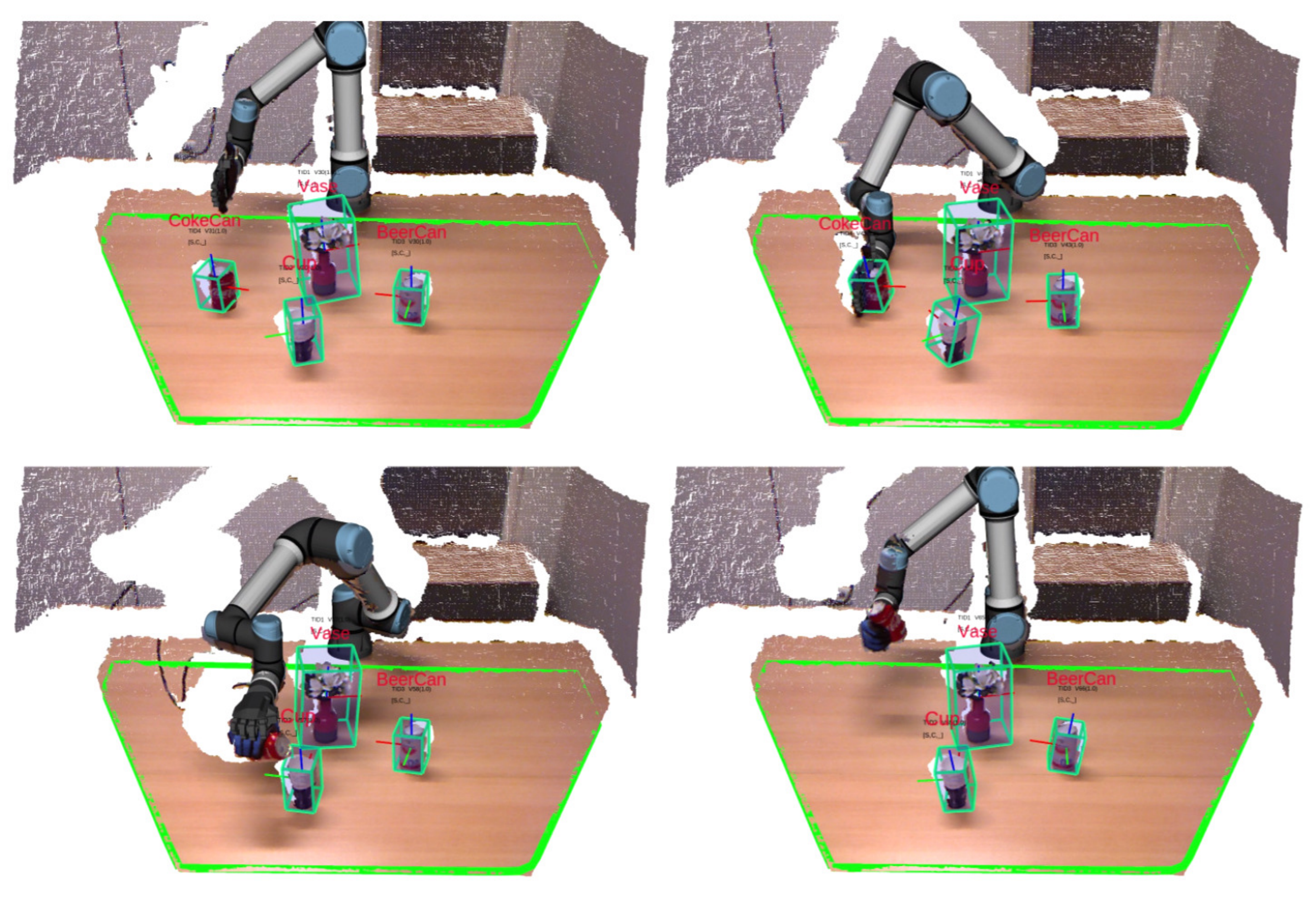}
\vspace{-4mm}
\caption{An example of object manipulation in \textit{serve\_a\_coke} scenario{, in which a robotic arm pours an opened CokeCan into a Cup}; (\textit{top-right}) Initially, the robot detects the table as shown by the green polygon. Then, all table-top objects are detected and recognized. The pose and category label of each object is highlighted by a bounding box and a label in red on top of each object. (\textit{top-left}) The robot then finds out the CokeCan object and goes to its pre-grasp area and picks it up first from the table. (\textit{bottom-left}) The robot retrieves the position of Cup first, and then calculates an obstacle-free trajectory and moves the CokeCan on top of the Cup to {serve} the drink. (\textit{bottom-right}) Finally, the robot computes an obstacle-free trajectory to navigate the robot’s end-effector to the initial pose.}
\label{serve_a_coke}
\end{figure}

Object manipulation is one of the challenging tasks in robotics and requires knowledge from different fields. In this section, we explain the details of object manipulation using an example; consider \textit{serve\_a\_coke} task as shown in Fig.~\ref{serve_a_coke}. To accomplish this task, the robot needs to detect and recognize all objects first. Afterwards, it has to identify a proper grasp pose for the coke object and plan an obstacle-free trajectory to reach the target pose. In such robotic task, an obstacle can be (\textit{i}) ``\textit{fixed objects}'' (e.g., wall, table, etc.), (\textit{ii}) objects that are in a fixed position most of the time (e.g., a \textit{vase}, a \textit{table-sign}, etc.), or (\textit{iii}) a \textit{dynamic object}, which usually corresponds to human's or the robot's body. Then, the robot executes the computed trajectory to grasp the target object. The robot computes an optimal trajectory from the current pose to the target pose to navigate the robot’s end-effector to a desired pose, which is on top of the cup object. It should be noted that the object manipulation module should check whether any part of the manipulator is at risk of colliding with itself or with any obstacles. Finally, the manipulation module sends out the action to the execution manager module. As evidenced by this example, collision avoidance is an essential problem in object manipulation that has to be addressed carefully to make the robot capable of performing safe interaction with human users and the environment to accomplish a task. {To compound the issue, if the robot in question is not stationary, considerations need to be made for object placement relative to the robot's movements. Towards addressing this point, a closed-loop visual servoing-based controller, in the form of eye-in-hand controller or eye-to-hand controller, is usually considered. These types of controllers are discussed in a two-part guide \cite{Chaumette2006,Chaumette2007} as well as a review paper by Kragic et. al. \cite{Kragic2003}.}


In the following subsections we first review recent collision avoidance and dexterity techniques, and then discuss object manipulation in human-robot shared environments in great details. 

\subsection{Self-collision avoidance}
 The standard way of dealing with \textit{self-collision avoidance} (SCA) is either by planning feasible paths under constraints, or to react when the robot almost collides with itself. A planned route is susceptible to interruptions and thus only works in static environments \cite{fromm2017self,kang2019sampling,pan2012collision}. The environments that service robots have to perform in, may not always be static. Whenever the environment changes, the plan will have to be recalculated, including the SCA. Reactive systems, such as \cite{rakita2018relaxedik,mirrazavi2018unified}, are more adaptable and can deal with changes to the environment. As a trade-off, they tend to get stuck more easily, may have trouble with tight spaces and tend to oscillate around local maxima in their joint space.

 Salehian et al., \cite{mirrazavi2018unified} proposed a new way of handling SCA. The collision boundary and its gradient in the combined joint space of the two arms is encoded into a model before the operation of the machine even starts. This stored gradient leads the robotic arms to avoid situations where they might collide, by being repulsed by the boundary denoting imminent collision. This distance to collision is encoded in a kernel {support vector machine (SVM)} and can be used as a constraint for IK solving instead of the usual joint-to-joint distances. This model is effectively data-driven, requiring examples of valid and invalid joint configurations for the SVM to learn. The trade-off to this is that solving the IK with SCA is very fast (\textless 10ms). The collision boundary and gradient have to be determined only once for a given robot blueprint and can be copied onto each production model.

While \cite{mirrazavi2018unified} has a good method to avoid collisions between its two robot arms, it does not take into account the singularities that can occur during motion. The model from \cite{rakita2018relaxedik} manages to balance SCA as well as avoiding singularities, moving smoothly and reaching the desired goal position and orientation. The SCA is done in a similar manner as before, but a neural network is trained instead of a SVM. The imminence of collisions, together with other objective functions, determines the constraints that the IK solver is under.

These methods are able to quickly avoid self-collision, but rely on data to show the collision boundaries. The safety of joint configurations can be easily simulated and so it is possible to generate this data beforehand. It is somewhat analogous to how humans learn the limits of their bodies, but without the need for some form of stress or touch sensors embedded into the robot arm structure. Nevertheless, looking into more human-like robotic arms, complete with additional sensors, can be useful for more advanced service robotics. However, for simpler service robots, the current implementations of collision proximity is more than enough.

\subsection{Avoiding collisions with the environment}

\begin{figure}[!t]
\includegraphics[width=\linewidth, trim= 10cm 5cm 0cm 0cm,clip=true]{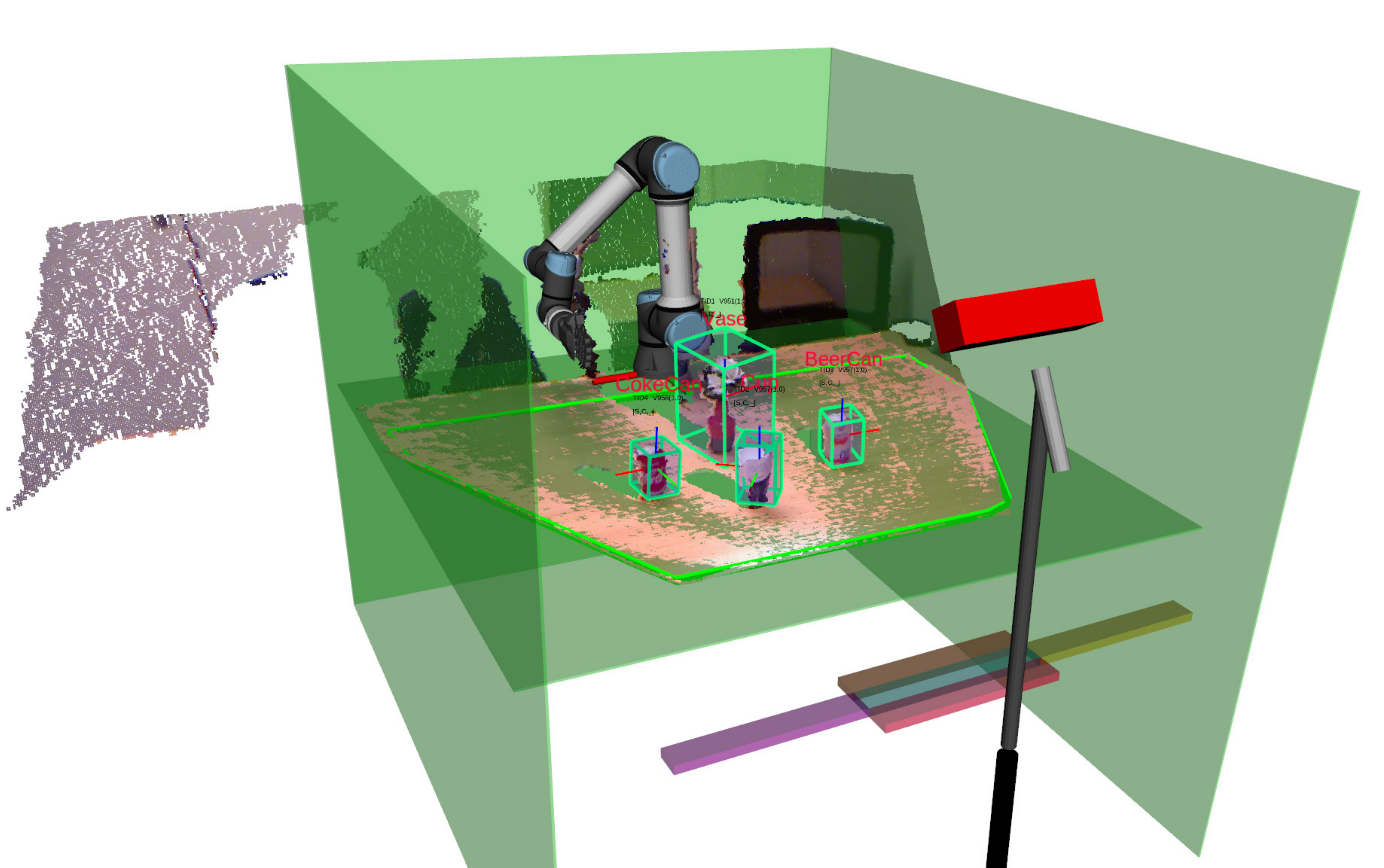}
\vspace{-4mm}
\caption{An example of defining a set of environmental constraints to prevent the robot from collisions with the environment in the \textit{serve\_a\_coke} scenario.}
\label{collision_with_enviroment}
\vspace{-2mm}
\end{figure}

In contrast to self-collision avoidance, checking for \textit{collisions with the environment} is a computationally intensive process (see Fig.~\ref{collision_with_enviroment}). Reactive systems have to continuously check if they are at risk of colliding while planners have to check every configuration or position that they may attempt to use. In an online environment, speed is key, an algorithm that takes more than a second may already be too slow. One way to remedy this in a planning-based approach is to sample more aggressively towards the goal position and orientation \cite{kang2019sampling}. This reduces the amount of nodes that have to be visited during planning, but also the amount of collision checks that have to be made. {Another approach is to consider continues time trajectories as being sampled from a Gaussian process \cite{mukadam2018continuous}. By using structure exploiting Gaussian process regression, the Gaussian Process Motion Planner can be used, which is an gradient based optimization algorithm. This algorithm can be made even more efficient by using factor graphs.}

Looking at the consequences of the actions is important, especially when it comes to hard hitting robots and delicate surroundings. One interesting way of preventing damage is by predicting the consequences of disturbing scenes \cite{fromm2017self}. When an object is picked up by the robot, it may cause other objects to move, tumble or fall. Objects may get damaged in this way, which is of course not desired. To remedy this, robots usually act as little as possible to get the job done. This new method instead focuses on learning the order in which to move objects to cause the least amount of damage. In this case, the paths that disturbed objects take, by rolling or falling, act as a cost or penalty. Based on available knowledge of the scene, the model is able to generalize and choose the best order of operations to cause the least damage. {Similarly~\cite{stilman2007manipulation} chooses to first move any blocking objects in the way of the goal object samples its actions based on a reverse-time search, which limits the number of objects that need to be moved aside.} 

{Other work focusing on motion planning in cluttered and dynamic environments has also been conducted. One promising approach proposed in~\cite{moll2017randomized} makes use of a physics engine to account for the uncertainty and dynamical interactions in such an environment. This has the benefit that explicit high level reasoning is no longer necessary for the desired manipulations. A limitation is that they focus on relatively simple grasping and manipulation, while physically far more advanced manipulations are common in every day life, think for example about throwing objects or using tools to manipulate other objects. To that end,~\cite{toussaint2018differentiable} introduce an integrated task and motion planning framework that includes the possibility of such advanced physical interactions, which combines continues path optimization with discrete logic for combinations of allowed interaction modes.}

{When the focus is more on efficient and simple manipulation,~\cite{kappler2018} {shows that} a tight coupling between real-time perception and reactive motion planning can lead to fast and accurate manipulation. Another, very cost effective approach has been proposed in~\cite{garrett2015ffrob}. They take the heuristic properties of the FastForward planner, a powerful symbolic planner, and apply them to motion planning. Combined with multi-query data structures, this leads to an efficient and integrated system for motion planning.}

Avoiding damage and reducing planning load are both important steps to get closer to fast and safe robotics. Nevertheless, not much work seems to be put into reducing the complexity of collision checking itself. For SCA, neural networks were trained to learn the boundaries of self collision. This resulted in a fast method of avoiding bad joint configurations. 
Such system could be set up to do a similar thing for visible objects with, for example, a stream point clouds as input. Similar to affordances for grasping, {humans seem to be able to quickly} ascertain affordances for arm and manipulator movement and orientation based on what they see and know. For service robots, this means they should be able to operate on a similar level, or at least fast enough to perform the various tasks we require of them. Towards this goal, Qureshi et al., proposed MPnet algorithm which is a learning-based neural planner for solving motion planning problems~\cite{qureshi2019motionICRA,qureshi2019motion}. In particular, they presented a recursive algorithm that receives environment information as point-clouds, as well as a robot's initial and desired goal configurations as input arguments and generates connectable path as an output. Some other researches also demonstrated segmenting motion tasks could reduce the complexity that the lower modules have to deal with~\cite{Sha2020}. Whether it is by distinguishing between individual tasks based on relations of touch \cite{aein2018library} or embedding scores or attractors directly into a robotic joint feature space \cite{Jetchev2014}. Simplifying the problem of motion with several degrees of freedom, allows for a greater focus on the other parts. {Using a voxel based model of the environment, efficient algorithms and data structures optimized for Graphics Processing Units allows for very efficient collision detection~\cite{hermann2014unified}.}

\subsection{Coordination and dexterity} 
\begin{figure}[!b]
\includegraphics[width=\linewidth, trim= 0cm 0cm 0cm 0cm,clip=true]{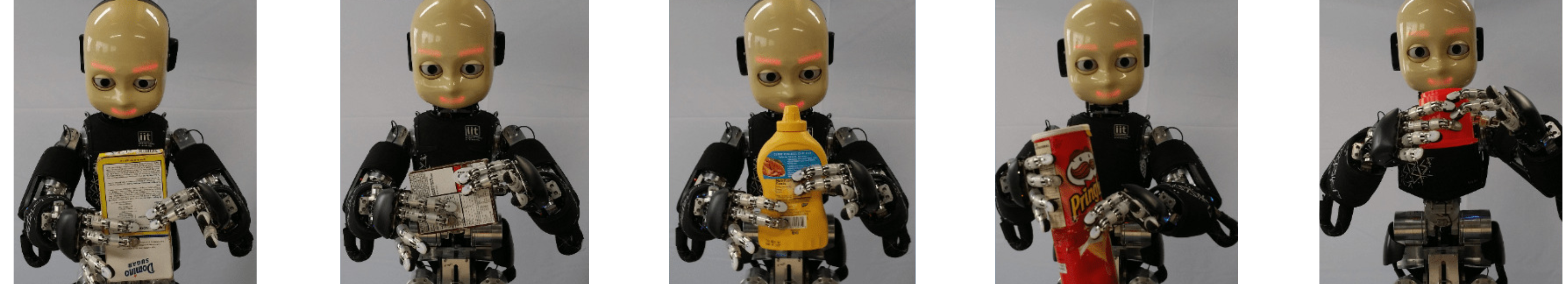}
\vspace{-2mm}
\caption{{A set of snapshots showing successful multi-hands grasping for various objects~\cite{vezzani2017novel}.}}
\label{handover_task}
\vspace{-2mm}
\end{figure}

For humans, doing something with ``one hand behind their back'' is seen as a challenge. Similarly for robots, doing a task with only one gripper is often possible, but may be more difficult. Furthermore, some tasks may even be impossible to perform. Coordinated and uncoordinated motion of two robotic arms is discussed in \cite{mirrazavi2018unified}, but in a factory setting instead of a household one. As household situations usually deal with objects of a smaller size, the self collision distance thresholds are set a bit higher than would be required. This highlights one of the potential problems with multi-arm coordination in household settings. Robotic hands may have to work very closely together, to the point of touching~\cite{vezzani2017novel} {(see Fig.~\ref{handover_task})}. As of yet, very little research seems to include this as a point of interest. While it is possible to manipulate objects one at a time in a cluttered environment \cite{zeng2018robotic} and many actions are possible with one arm, there are cases where extra dexterity is required, {mainly in social human-robot collaborative scenarios \cite{mirrazavi2018unified,busch2019evaluation}.}

{When it comes to advanced manipulation with one hand, there is also still room for improvement. An extensive review of such advanced manipulations and current limitations is provided in~\cite{billard2019trends}, which gives an overview of all aspects involved, including hardware requirements, sensory systems, learning algorithms and interaction with humans. The most difficult part of these manipulations is usually not the grasping itself, but the additional manipulation that is necessary when the object is already grasped.} Preventing an object from slipping from a gripper \cite{huang2017intelligent} and minor manipulation with fingers while grasping \cite{sundaralingam2019relaxed} have already been somewhat explored. These all lay the foundation for advanced manipulation, which remains mostly out of focus in favor of the act of grasping itself. Nevertheless, advanced manipulation is a requirement for the most delicate of tasks. As an example, correctly cracking open an egg without any tools is difficult even for humans, but a service robot may at one point be asked to do this or similar difficult tasks. For this reason, advanced manipulation still requires more research.

\subsection{Manipulation in human-robot shared environments} 
\label{manipulation_in_human_robot_shared_environments} 

A lot of research has gone into enabling robots to deal with a wide variety of environments, both static and dynamic. While static environments have been mastered quite well, the navigation of dynamic environments, especially those that deal with having other human agents in the same space as the robot, are currently a hot topic in the field. Path and trajectory planning are essential components of object manipulation, especially if the robot is required to affect the environment past its immediate reach. This means that it must be able to relocate itself in, at worst, a highly dynamic environment with one or more other agents that may act unpredictably. We can further break down this problem into local and global navigation. Global navigation deals with planning towards a goal or objective that is not currently in the range of the robot's perception while local navigation concerns navigation through the immediately perceivable space around the robot~\cite{chaves2019integration}. Global navigation and path planning has been largely tackled to a satisfactory degree and should generally be able to converge to an optimal global solution \cite{karaman2011globalnav}. 

One of the focuses in current developments is on local path planning and obstacle avoidance in dynamic environments. Within this task, we can further specialize into varying levels of application. Lower level concepts, such as combining color data with depth sensing to get additional information about the environment~\cite{cherubini2013visual,cao2018openpose}, can be used for collision avoidance. A slightly higher level task is creating a collision risk map in order to more effectively navigate the environment. These risk maps provide some insight into possible future states from the current one~\cite{chaves2019integration,liang2016intelligent}. Instead of explicitly modelling the collision risk using an artificial neural network, one can instead be trained to abstract the risk to higher or lower confidence path solutions \cite{singh2019neural}.

When considering a robot's movement in human-robot shared environments, many of the lower level functions and considerations have been puzzled out already, so the current state-of-the-art instead focuses on higher level control philosophies. For a truly robust trajectory planning in such a dynamic environment, the \textit{socially aware control model} should be able to account for unexpected events, such as groups or individuals moving away from or avoiding unmodeled obstacles. Expanding on \cite{zeng2013mobile} and the concept of prioritizing human agents in an environment, we can see a shift towards increasingly social based models, incorporating social force models (SFM) \cite{helbing1995sfm} to develop a system of socially reactive control (SRC) \cite{truong2017socially}. Such a system proposes to take into account not only single humans, but groups of them and their collective motion for mobile manipulators, which can be used in stationary manipulation scenarios as well. This means group dynamics, such as group motion, centres and size of groups. Additionally it is proposed to gain some basic understanding of what the human agent is doing at a given time, what they are interacting with, and using that information to further understand what is likely to happen in the environment around the robot. 

An example given in the paper \cite{truong2017socially} shows a human interacting with an object of interest and a second human facing the first as if to approach them. This would cause either the robot to cross the path of the human or the human to cross the robot's path if both continue on a straight line. Additionally, in the path to the goal are two humans who clearly form another distinct group. The socially reactive model of control will attempt to avoid the group as a whole instead of attempting to pass through them, even if there is enough space to perform such an action. While group dynamics can be useful to model in order to avoid collision or interactions with groups of humans, the model of navigation, considering humans as single independent agents rather than trying to infer groups, is still very relevant. In this case, the state-of-the-art proposes a change from a flat pre-calculated confidence for the trajectory of a human to a Bayesian one \cite{fridovich2019confidence} instead, which is constantly updated. This would allow for unobtrusive navigation around agents that act entirely unpredictable. This means either intrinsically unpredictable or unpredictable in a sense that an agent is reacting to features of the environment that are not modelled by the robot, which would make a subsequent action to avoid an obstacle or object unpredictable. Such a motion plan is successful at avoiding human agents because it is far more conservative in the planning stage, considering a much wider area around a human to be inaccessible. In essence, it is similar to the earlier introduced concept of a virtual ``\textit{force-field}'' around a human agent \cite{zeng2013mobile}, however it is not modelled explicitly as such. This provides a slightly more adaptable framework where any other agent in motion, human or not, is able to be avoided, by not making too strong of an assumption as to its intended direction of motion or goal.

Another interesting thing to consider in human-robot shared environments is that all paths to a certain goal may be blocked, even by very light or easily movable objects. Should a robot, in its navigation, consider if an object can be pushed aside without damage to itself or the environment? For example, a sheet of cardboard slips from a shelf and blocks the robot's path, can the robot simply push through it as it would do no damage to either the robot or the rest of the environment? Navigation has been considered as a tool to aid in a robot's vision and perception, especially in crowded scenarios, and generally serves the purpose of moving the robot to a location where it can manipulate the environment using its gripper or other manipulation tool. However, \textit{should the navigation itself be considered a valid method for manipulating the environment, using the frame of the robot itself?} {This approach has been considered as a valid method of moving through an environment and affecting it~\cite{wilfong1991motion}. In this paper, the argument is made for considering an environment where the agent encounters movable obstacles. This consideration applies to both navigational movements and movements with a manipulator however{,} as stated previously, these can be considered one and the same for these purposes. Objects have either a modelled path and goal or one that is completely unknown. This creates problems for planning as the problem immediately becomes very computationally intense in 3 dimensional space. For this, Wilfong proposes a branching method of exploring the potential future states of the environment. Additionally, an expansion to these methods is an approach {modelling} potential future states with attached probabilities of {reaching} them in order to eliminate unnecessary complexity~\cite{van2009path}.}

Another concept to consider, is something more akin to mapping. Whereas so far, we have discussed path and trajectory planning (i.e., navigation) as a means to bring a manipulator to a desired location in a dynamic environment, we can instead consider another intermediary task. The camera of a robot rarely has more than just the three rotational degrees of freedom, so any planar motion that is desired needs to be provided by the robot itself through navigation. Yervilla-Herrera et al., \cite{yervilla2019optimal} proposed a use-case for navigation which combines some basic static obstacle avoidance with the principles of object reconstruction using methods like \textit{shape from motion}. In this case, the goal of the navigation is, in fact, to provide better or more complete sensory information to the robot, rather than navigating to a specific point in the space. This may be useful in the case that an object in a pile is more easily detected or manipulated from a different angle of approach \cite{kasaei2018perceiving,sock2017multi}, or simply to gain a better understanding of the overall shape of an object. As mentioned in section~\ref{perception}, the better the robot's ability to perceive an object, the easier it is to manipulate {that object}. 

\section{Conclusion and Future Work}
\label{conclusion} 

In recent years, many great developments have been made in the field of service robotics. 
It can be seen that a lot of recent developments are due to the use of more complex machine learning techniques, such as (deep) neural networks, and are based on large amounts of data. While this leads to continuous improvements on the tasks themselves, large issues remain with real-world applicability due to time complexity issues. Service robots also struggle severely in unknown environments, lacking open-ended learning about object categories and scenes, a map for global navigation, reliable object recognition for local navigation, and running into collision issues while manipulating objects. After these robotic tasks are solved in experimental setups, the focus will need to shift from solutions, to applicability, by reducing complexity and implementing more open-ended learning techniques.

As reviewed in this paper, several major issues and hurdles are solved almost entirely. It is shown that path planning is close to completion in a reliable, known environment, as is grasp planning. Similarly, object recognition, which feeds into both these types of planning, is also approaching near perfect scores when applied to objects in a \textit{predetermined setting}.

Current issues in object manipulation can be boiled down to avoiding collision. This means mostly dealing with dynamic and shared human-robot environments by developing better methods for local planning. Some solutions that have been developed recently concern themselves with humans, or even groups of humans. Using group dynamics, a group of humans or other agents can temporarily be seen as a single unit for the purposes of trajectory prediction. Whether tracking a single human or a group, trajectory prediction has also undergone development {through} Bayesian trajectory calculations. This allows the robot a range of possibilities where an entity might move next and use this probability to plan its own path. Besides, self-collision avoidance is also addressed in recent works. New approaches have been suggested which are more adaptable and can adjust a grasp movement during execution. These adaptive approaches have a higher tendency to get stuck, however. Through the use of SVM~\cite{mirrazavi2018unified}, resulting in adaptable grasps that sometimes ran into singularities, and later using a neural net for motion planning~\cite{qureshi2019motion}, these issues were resolved.

\subsection{Trade-offs}
While some areas see considerable linear improvements, other areas suffer from one of two situations: either an improvement in one criteria, such as accuracy, proves to be a setback in another criteria, such as speed, or two approaches are developed side-by-side and their development continuously surpasses each other, without a clear best approach to the task.

The first category is mainly concerning planning modules. When a planning module becomes more sophisticated, usually, the complexity of the constraints increases. This causes the calculations to be much more difficult, almost inevitably affecting the time it takes to find a proper solution. As such, in both navigation and grasp planning, continuous issues arise when trying to apply new methods in real environments. While some of these computational burdens eventually even out due to increases in hardware capacity, other times specific research needs to be done to reduce time complexity.

The second category is seen across many different areas of service robotics. In object perception, it applies to object representation, differentiating between object descriptors that are either hand-crafted or trained by a neural network. It can also be seen in how to view the environment, where approaches based on bounding boxes compete with approaches using image segmentation. Finally, in more recent work on object perception, the problem of cluttered areas is addressed, where multiple objects are in a pile. Here the distinction is made between using a grasping module to relocate the items before identifying them, or to use segmentation to label the partially occluded objects~\cite{eitel2020learning}.

In grasping, certain trade-offs from this second category can also be observed. With the problem of damage reduction due to collision, some approaches try to limit the movements made by a manipulator such that the chance of an accident is limited. Other, more dynamic solutions{,} will try to predict {possible} damaging results and use this damage as a parameter or constraint in grasp planning. Finally, when constructing the grasp movements, there is no consensus on whether grasps should be based on point clouds, segmented by a rule-based system, or by using neural nets to generate grasps. From this list of trade-offs, it can be seen that the field of robotics delivers far from a unified solution to most issues. New approaches are continuously developed, old approaches are reinvigorated and improved and even opposing, but similarly effective approaches are found.

\subsection{Future Work}
While new developments are made frequently, some issues are either not solved or largely lacking research. For object manipulation, a suggested direction is to focus on different control philosophies for local planning. While some, such as Bayesian trajectory prediction, are already being developed, this is the area where navigation has most to gain. Some suggest that navigation can also be used to have a robot capable of learning online, basically using navigation to explore. This would mean a robot needs to determine the most probable locations for a given object in a household environment or part of a map, and use navigation planning to go to that locations to observe and manipulate the object.

A different issue remaining for the grasping task is that of increasing complexity and dependency on data. Many models are currently trained on large sets of data from previous grasps, or on data concerning specific objects. This means that open-ended learning, as with the object perception task, is still lacking. To the best of the authors' knowledge, object detection for manipulation has largely been used for small tabletop items. However, the robot should also be able to autonomously manipulate larger objects like (wheel) chairs, as well as partly fixed objects like cabinet doors or windows. To this end, it is important to appropriately extend 3D object recognition and affordance prediction to furniture, doors and windows.

For object perception, the list of unresolved issues is longer. It contains, among other things, dealing with objects with reflective surfaces and dealing with large objects. In order to solve the more difficult corner cases, several suggestions have been made. One such suggestion is to incorporate other forms of perception, such as tactile, into object recognition. In order to be less data dependent, the area of object perception will need to put a bigger emphasis on open-ended learning, allowing a robot to learn new objects while performing tasks. The final addition to object perception overlaps with grasping{,} as it has to do with affordance predictions. Affordances have been used in both perception and grasping for the identification of objects. As an additional use, grasping affordances can be extracted from images to quickly find suitable grasping locations and orientations. It may be possible to extract other types as affordances such as ease of movement for terrain, available space to move in 3D or joint spaces. As affordances are already used for grasping with single grippers, they may be applied to the case of double-armed robots as well. As little work has been done on the use of multiple grippers simultaneously in the household, this seems like a worthwhile direction.



%
%

\bibliographystyle{spmpsci}      
{\small
\bibliography{literature}}   

\begin{thebibliography}{100}
\providecommand{\url}[1]{{#1}}
\providecommand{\urlprefix}{URL }
\expandafter\ifx\csname urlstyle\endcsname\relax
  \providecommand{\doi}[1]{DOI~\discretionary{}{}{}#1}\else
  \providecommand{\doi}{DOI~\discretionary{}{}{}\begingroup
  \urlstyle{rm}\Url}\fi

\bibitem{bostondynamicsSpot}
Boston dynamics spot.
\newblock \urlprefix\url{https://www.bostondynamics.com/spot}

\bibitem{shadowrobotcompany}
Dexterous hand.
\newblock \urlprefix\url{https://www.shadowrobot.com/products/dexterous-hand/}

\bibitem{husqvarna}
husqvarna automower.
\newblock
  \urlprefix\url{https://www.husqvarna.com/us/products/robotic-lawn-mowers/}

\bibitem{iRobot}
irobot.
\newblock \urlprefix\url{https://www.irobot.com/}

\bibitem{aein2018library}
Aein, M.J., Aksoy, E.E., W{\"o}rg{\"o}tter, F.: Library of actions:
  Implementing a generic robot execution framework by using manipulation action
  semantics.
\newblock The International Journal of Robotics Research p. 910–934 (2018)

\bibitem{grapeNose}
Aleixandre, M., Santos, J.P., Sayago, I., Cabellos, J.M., Arroyo, T., Horrillo,
  M.C.: A wireless and portable electronic nose to differentiate musts of
  different ripeness degree and grape varieties.
\newblock Sensors \textbf{15}(4), 8429--8443 (2015)

\bibitem{antanas2019semantic}
Antanas, L., Moreno, P., Neumann, M., de~Figueiredo, R.P., Kersting, K.,
  Santos-Victor, J., De~Raedt, L.: Semantic and geometric reasoning for robotic
  grasping: a probabilistic logic approach.
\newblock Autonomous Robots \textbf{43}(6), 1393--1418 (2019)

\bibitem{rightDirectionSmell}
Arain, M.A., Schaffernicht, E., Bennetts, V.H., Lilienthal, A.J.: The right
  direction to smell: Efficient sensor planning strategies for robot assisted
  gas tomography.
\newblock In: 2016 IEEE International Conference on Robotics and Automation
  (ICRA), pp. 4275--4281. IEEE (2016)

\bibitem{davinci}
Arumugam, R., Enti, V.R., Bingbing, L., Xiaojun, W., Baskaran, K., Kong, F.F.,
  Kumar, A.S., Meng, K.D., Kit, G.W.: Davinci: A cloud computing framework for
  service robots.
\newblock In: 2010 IEEE international conference on robotics and automation,
  pp. 3084--3089. IEEE (2010)

\bibitem{ARMAR}
Asfour, T., W\"achter, M., Kaul, L., Rader, S., Weiner, P., Ottenhaus, S.,
  Grimm, R., Zhou, Y., Grotz, M., Paus, F.: {ARMAR}-6: A high-performance
  humanoid for human-robot collaboration in real world scenarios.
\newblock IEEE Robotics \& Automation Magazine \textbf{26}(4), 108--121 (2019)

\bibitem{Rosie}
Beetz, M., Klank, U., Kresse, I., Maldonado, A., M{\"o}senlechner, L.,
  Pangercic, D., R{\"u}hr, T., Tenorth, M.: Robotic roommates making pancakes.
\newblock In: 2011 11th IEEE-RAS International Conference on Humanoid Robots,
  pp. 529--536. IEEE (2011)

\bibitem{cloudPerception1}
Beksi, W.J., Spruth, J., Papanikolopoulos, N.: Core: A cloud-based object
  recognition engine for robotics.
\newblock In: 2015 IEEE/RSJ International Conference on Intelligent Robots and
  Systems (IROS), pp. 4512--4517. IEEE (2015)

\bibitem{billard2019trends}
Billard, A., Kragic, D.: Trends and challenges in robot manipulation.
\newblock Science \textbf{364}(6446) (2019)

\bibitem{bochkovskiy2020yolov4}
Bochkovskiy, A., Wang, C.Y., Liao, H.Y.M.: Yolov4: Optimal speed and accuracy
  of object detection.
\newblock arXiv preprint arXiv:2004.10934  (2020)

\bibitem{bohg2013data}
Bohg, J., Morales, A., Asfour, T., Kragic, D.: Data-driven grasp synthesis—a
  survey.
\newblock IEEE Transactions on Robotics \textbf{30}(2), 289--309 (2013)

\bibitem{busch2019evaluation}
Busch, B., Cotugno, G., Khoramshahi, M., Skaltsas, G., Turchi, D., Urbano, L.,
  W{\"a}chter, M., Zhou, Y., Asfour, T., Deacon, G., et~al.: Evaluation of an
  industrial robotic assistant in an ecological environment.
\newblock In: 2019 28th IEEE International Conference on Robot and Human
  Interactive Communication (RO-MAN), pp. 1--8. IEEE (2019)

\bibitem{cao2018openpose}
Cao, Z., Hidalgo, G., Simon, T., Wei, S.E., Sheikh, Y.: Openpose: realtime
  multi-person 2d pose estimation using part affinity fields.
\newblock IEEE transactions on pattern analysis and machine intelligence
  \textbf{43}(1), 172--186 (2019)

\bibitem{Distance}
Cha, S.H.: Comprehensive survey on distance/similarity measures between
  probability density functions.
\newblock City \textbf{1}(2), 1 (2007)

\bibitem{chang2012interactive}
Chang, L., Smith, J.R., Fox, D.: Interactive singulation of objects from a
  pile.
\newblock In: 2012 IEEE International Conference on Robotics and Automation,
  pp. 3875--3882. IEEE (2012)

\bibitem{Chaumette2006}
Chaumette, F., Hutchinson, S.: Visual servo control. i. basic approaches.
\newblock IEEE Robotics {\&}amp; Automation Magazine \textbf{13}(4) (2006).
\newblock \doi{10.1109/MRA.2006.250573}.
\newblock \urlprefix\url{https://doi.org/10.1109/MRA.2006.250573}

\bibitem{Chaumette2007}
Chaumette, F., Hutchinson, S.: Visual servo control. ii. advanced approaches
  [tutorial].
\newblock IEEE Robotics {\&}amp; Automation Magazine \textbf{14}(1) (2007).
\newblock \doi{10.1109/MRA.2007.339609}.
\newblock \urlprefix\url{https://doi.org/10.1109/MRA.2007.339609}

\bibitem{chaves2019integration}
Chaves, D., Ruiz-Sarmiento, J., Petkov, N., Gonzalez-Jimenez, J.: Integration
  of {CNN} into a robotic architecture to build semantic maps of indoor
  environments.
\newblock In: International Work-Conference on Artificial Neural Networks, pp.
  313--324. Springer (2019)

\bibitem{deepSegmentation}
Chen, L.C., Papandreou, G., Kokkinos, I., Murphy, K., Yuille, A.L.: {DeepLab}:
  Semantic image segmentation with deep convolutional nets, atrous convolution,
  and fully connected {CRFs}.
\newblock IEEE transactions on pattern analysis and machine intelligence
  \textbf{40}(4), 834--848 (2017)

\bibitem{cherubini2013visual}
Cherubini, A., Chaumette, F.: Visual navigation of a mobile robot with
  laser-based collision avoidance.
\newblock The International Journal of Robotics Research \textbf{32}(2),
  189--205 (2013)

\bibitem{NoseIndustry}
Chilo, J., Pelegri-Sebastia, J., Cupane, M., Sogorb, T.: E-nose application to
  food industry production.
\newblock IEEE Instrumentation \& Measurement Magazine \textbf{19}(1), 27--33
  (2016)

\bibitem{choy20194d}
Choy, C., Gwak, J., Savarese, S.: {4D} spatio-temporal convnets: Minkowski
  convolutional neural networks.
\newblock In: Proceedings of the IEEE Conference on Computer Vision and Pattern
  Recognition, pp. 3075--3084 (2019)

\bibitem{fingerTaste}
Ciui, B., Martin, A., Mishra, R.K., Nakagawa, T., Dawkins, T.J., Lyu, M.,
  Cristea, C., Sandulescu, R., Wang, J.: Chemical sensing at the robot
  fingertips: Toward automated taste discrimination in food samples.
\newblock ACS sensors \textbf{3}(11), 2375--2384 (2018)

\bibitem{debeunne2020review}
Debeunne, C., Vivet, D.: A review of visual-lidar fusion based simultaneous
  localization and mapping.
\newblock Sensors \textbf{20}(7), 2068 (2020)

\bibitem{deimel2016novel}
Deimel, R., Brock, O.: A novel type of compliant and underactuated robotic hand
  for dexterous grasping.
\newblock The International Journal of Robotics Research \textbf{35}(1-3),
  161--185 (2016)

\bibitem{do2018affordancenet}
Do, T.T., Nguyen, A., Reid, I.: {AffordanceNet}: An end-to-end deep learning
  approach for object affordance detection.
\newblock In: 2018 IEEE international conference on robotics and automation
  (ICRA), pp. 1--5. IEEE (2018)

\bibitem{doumanoglou2016recovering}
Doumanoglou, A., Kouskouridas, R., Malassiotis, S., Kim, T.K.: Recovering {6D}
  object pose and predicting next-best-view in the crowd.
\newblock In: Proceedings of the IEEE Conference on Computer Vision and Pattern
  Recognition, pp. 3583--3592 (2016)

\bibitem{eckert2008cross}
Eckert, M.A., Kamdar, N.V., Chang, C.E., Beckmann, C.F., Greicius, M.D., Menon,
  V.: A cross-modal system linking primary auditory and visual cortices:
  Evidence from intrinsic {fMRI} connectivity analysis.
\newblock Human brain mapping \textbf{29}(7), 848--857 (2008)

\bibitem{eitel2020learning}
Eitel, A., Hauff, N., Burgard, W.: Learning to singulate objects using a push
  proposal network.
\newblock In: Robotics Research, pp. 405--419. Springer (2020)

\bibitem{elango2015review}
Elango, N., Faudzi, A.: A review article: investigations on soft materials for
  soft robot manipulations.
\newblock The International Journal of Advanced Manufacturing Technology
  \textbf{80}(5-8), 1027--1037 (2015)

\bibitem{TORO}
Englsberger, J., Werner, A., Ott, C., Henze, B., Roa, M.A., Garofalo, G.,
  Burger, R., Beyer, A., Eiberger, O., Schmid, K., et~al.: Overview of the
  torque-controlled humanoid robot {TORO}.
\newblock In: 2014 IEEE-RAS International Conference on Humanoid Robots, pp.
  916--923. IEEE (2014)

\bibitem{ephrat2018looking}
Ephrat, A., Mosseri, I., Lang, O., Dekel, T., Wilson, K., Hassidim, A.,
  Freeman, W.T., Rubinstein, M.: Looking to listen at the cocktail party: A
  speaker-independent audio-visual model for speech separation.
\newblock arXiv preprint arXiv:1804.03619  (2018)

\bibitem{eppner2015planning}
Eppner, C., Brock, O.: Planning grasp strategies that exploit environmental
  constraints.
\newblock In: 2015 IEEE International Conference on Robotics and Automation
  (ICRA), pp. 4947--4952. IEEE (2015)

\bibitem{ernst2004merging}
Ernst, M.O., B{\"u}lthoff, H.H.: Merging the senses into a robust percept.
\newblock Trends in cognitive sciences \textbf{8}(4), 162--169 (2004)

\bibitem{evans2008dual}
Evans, J.S.B.: Dual-processing accounts of reasoning, judgment, and social
  cognition.
\newblock Annu. Rev. Psychol. \textbf{59}, 255--278 (2008)

\bibitem{faulhammer2016autonomous}
F{\"a}ulhammer, T., Ambru{\c{s}}, R., Burbridge, C., Zillich, M., Folkesson,
  J., Hawes, N., Jensfelt, P., Vincze, M.: Autonomous learning of object models
  on a mobile robot.
\newblock IEEE Robotics and Automation Letters \textbf{2}(1), 26--33 (2016)

\bibitem{ficuciello2019hand}
Ficuciello, F.: Hand-arm autonomous grasping: Synergistic motions to enhance
  the learning process.
\newblock Intelligent Service Robotics \textbf{12}(1), 17--25 (2019)

\bibitem{fontanals2014integrated}
Fontanals, J., Dang-Vu, B.A., Porges, O., Rosell, J., Roa, M.A.: Integrated
  grasp and motion planning using independent contact regions.
\newblock In: 2014 IEEE-RAS International Conference on Humanoid Robots, pp.
  887--893. IEEE (2014)

\bibitem{fridovich2019confidence}
Fridovich-Keil, D., Bajcsy, A., Fisac, J.F., Herbert, S.L., Wang, S., Dragan,
  A.D., Tomlin, C.J.: Confidence-aware motion prediction for real-time
  collision avoidance.
\newblock The International Journal of Robotics Research pp. 250--265 (2019)

\bibitem{fromm2017self}
Fromm, T.: Self-supervised damage-avoiding manipulation strategy optimization
  via mental simulation.
\newblock arXiv preprint arXiv:1712.07452  (2017)

\bibitem{DLR}
Fuchs, M., Borst, C., Giordano, P.R., Baumann, A., Kraemer, E., Langwald, J.,
  Gruber, R., Seitz, N., Plank, G., Kunze, K., et~al.: {Rollin'Justin-Design}
  considerations and realization of a mobile platform for a humanoid upper
  body.
\newblock In: 2009 IEEE International Conference on Robotics and Automation,
  pp. 4131--4137. IEEE (2009)

\bibitem{funkhouser2019robotic}
Funkhouser, T.A.: Robotic pick-and-place of novel objects in clutter with
  multi-affordance grasping and cross-domain image matching.
\newblock International Journal of Robotics Research  (2019)

\bibitem{gao2018learning}
Gao, R., Feris, R., Grauman, K.: Learning to separate object sounds by watching
  unlabeled video.
\newblock In: Proceedings of the European Conference on Computer Vision (ECCV),
  pp. 35--53 (2018)

\bibitem{2.5dVisualSound}
Gao, R., Grauman, K.: {2.5D} visual sound.
\newblock In: Proceedings of the IEEE Conference on Computer Vision and Pattern
  Recognition, pp. 324--333 (2019)

\bibitem{gao2019co}
Gao, R., Grauman, K.: Co-separating sounds of visual objects.
\newblock In: Proceedings of the IEEE International Conference on Computer
  Vision, pp. 3879--3888 (2019)

\bibitem{gao2013learning}
Gao, S., Tsang, I.W.H., Ma, Y.: Learning category-specific dictionary and
  shared dictionary for fine-grained image categorization.
\newblock IEEE Transactions on Image Processing \textbf{23}(2), 623--634 (2013)

\bibitem{garrett2015ffrob}
Garrett, C.R., Lozano-P{\'e}rez, T., Kaelbling, L.P.: Ffrob: An efficient
  heuristic for task and motion planning.
\newblock In: Algorithmic Foundations of Robotics XI, pp. 179--195. Springer
  (2015)

\bibitem{gidaris2018dynamic}
Gidaris, S., Komodakis, N.: Dynamic few-shot visual learning without
  forgetting.
\newblock In: Proceedings of the IEEE Conference on Computer Vision and Pattern
  Recognition, pp. 4367--4375 (2018)

\bibitem{fastRCNN}
Girshick, R.: Fast r-cnn.
\newblock In: Proceedings of the IEEE international conference on computer
  vision, pp. 1440--1448 (2015)

\bibitem{RCNN}
Girshick, R., Donahue, J., Darrell, T., Malik, J.: Rich feature hierarchies for
  accurate object detection and semantic segmentation.
\newblock In: Proceedings of the IEEE conference on computer vision and pattern
  recognition, pp. 580--587 (2014)

\bibitem{gupta2012using}
Gupta, M., Sukhatme, G.S.: Using manipulation primitives for brick sorting in
  clutter.
\newblock In: 2012 IEEE International Conference on Robotics and Automation,
  pp. 3883--3889. IEEE (2012)

\bibitem{grapeTongue}
Guti{\'e}rrez, M., Llobera, A., Ipatov, A., Vila-Planas, J., M{\'\i}nguez, S.,
  Demming, S., B{\"u}ttgenbach, S., Capdevila, F., Domingo, C.,
  Jim{\'e}nez-Jorquera, C.: Application of an e-tongue to the analysis of
  monovarietal and blends of white wines.
\newblock Sensors \textbf{11}(5), 4840--4857 (2011)

\bibitem{hang2016hierarchical}
Hang, K., Li, M., Stork, J.A., Bekiroglu, Y., Pokorny, F.T., Billard, A.,
  Kragic, D.: Hierarchical fingertip space: A unified framework for grasp
  planning and in-hand grasp adaptation.
\newblock IEEE Transactions on robotics \textbf{32}(4), 960--972 (2016)

\bibitem{hang2019pre}
Hang, K., Morgan, A.S., Dollar, A.M.: Pre-grasp sliding manipulation of thin
  objects using soft, compliant, or underactuated hands.
\newblock IEEE Robotics and Automation Letters \textbf{4}(2), 662--669 (2019)

\bibitem{hariharan2017low}
Hariharan, B., Girshick, R.: Low-shot visual recognition by shrinking and
  hallucinating features.
\newblock In: Proceedings of the IEEE International Conference on Computer
  Vision, pp. 3018--3027 (2017)

\bibitem{harnad2017cognize}
Harnad, S.: To cognize is to categorize: Cognition is categorization.
\newblock In: Handbook of categorization in cognitive science, pp. 21--54.
  Elsevier (2017)

\bibitem{haustein2017integrating}
Haustein, J.A., Hang, K., Kragic, D.: Integrating motion and hierarchical
  fingertip grasp planning.
\newblock In: 2017 IEEE International Conference on Robotics and Automation
  (ICRA), pp. 3439--3446. IEEE (2017)

\bibitem{maskRCNN}
He, K., Gkioxari, G., Doll{\'a}r, P., Girshick, R.: Mask r-cnn.
\newblock In: Proceedings of the IEEE international conference on computer
  vision, pp. 2961--2969 (2017)

\bibitem{helbing1995sfm}
Helbing, D., Moln\'ar, P.: Social force model for pedestrian dynamics.
\newblock Phys. Rev. E \textbf{51}, 4282--4286 (1995).
\newblock \doi{10.1103/PhysRevE.51.4282}.
\newblock \urlprefix\url{https://link.aps.org/doi/10.1103/PhysRevE.51.4282}

\bibitem{hermann2014unified}
Hermann, A., Drews, F., Bauer, J., Klemm, S., Roennau, A., Dillmann, R.:
  Unified gpu voxel collision detection for mobile manipulation planning.
\newblock In: 2014 IEEE/RSJ International Conference on Intelligent Robots and
  Systems, pp. 4154--4160. IEEE (2014)

\bibitem{hermans2011affordance}
Hermans, T., Rehg, J.M., Bobick, A.: Affordance prediction via learned object
  attributes.
\newblock In: ICRA: Workshop on Semantic Perception, Mapping, and Exploration,
  vol.~1. Citeseer (2011)

\bibitem{RACE}
Hertzberg, J., Zhang, J., Zhang, L., Rockel, S., Neumann, B., Lehmann, J.,
  Dubba, K.S., Cohn, A.G., Saffiotti, A., Pecora, F., et~al.: The {RACE}
  project.
\newblock KI-K{\"u}nstliche Intelligenz \textbf{28}(4), 297--304 (2014)

\bibitem{herzog2014learning}
Herzog, A., Pastor, P., Kalakrishnan, M., Righetti, L., Bohg, J., Asfour, T.,
  Schaal, S.: Learning of grasp selection based on shape-templates.
\newblock Autonomous Robots \textbf{36}(1-2), 51--65 (2014)

\bibitem{huang2017intelligent}
Huang, S.J., Chang, W.H., Su, J.Y.: Intelligent robotic gripper with adaptive
  grasping force.
\newblock International Journal of Control, Automation and Systems
  \textbf{15}(5), 2272--2282 (2017)

\bibitem{maskScoreRCNN}
Huang, Z., Huang, L., Gong, Y., Huang, C., Wang, X.: Mask scoring r-cnn.
\newblock In: Proceedings of the IEEE conference on computer vision and pattern
  recognition, pp. 6409--6418 (2019)

\bibitem{illeris2018comprehensive}
Illeris, K.: A comprehensive understanding of human learning.
\newblock In: Contemporary theories of learning, pp. 1--14. Routledge (2018)

\bibitem{ingrand2017deliberation}
Ingrand, F., Ghallab, M.: Deliberation for autonomous robots: A survey.
\newblock Artificial Intelligence \textbf{247}, 10--44 (2017)

\bibitem{smellReview}
Ishida, H., Wada, Y., Matsukura, H.: Chemical sensing in robotic applications:
  A review.
\newblock IEEE Sensors Journal \textbf{12}(11), 3163--3173 (2012)

\bibitem{El-E}
Jain, A., Kemp, C.C.: {EL-E}: an assistive mobile manipulator that autonomously
  fetches objects from flat surfaces.
\newblock Autonomous Robots \textbf{28}(1), 45 (2010)

\bibitem{Jetchev2014}
Jetchev, N., Toussaint, M.: Discovering relevant task spaces using inverse
  feedback control.
\newblock Autonomous Robots \textbf{37}(2), 169--189 (2014).
\newblock \doi{10.1007/s10514-014-9384-1}.
\newblock \urlprefix\url{https://doi.org/10.1007/s10514-014-9384-1}

\bibitem{johns2016deep}
Johns, E., Leutenegger, S., Davison, A.J.: Deep learning a grasp function for
  grasping under gripper pose uncertainty.
\newblock In: 2016 IEEE/RSJ International Conference on Intelligent Robots and
  Systems (IROS), pp. 4461--4468. IEEE (2016)

\bibitem{johnson1999using}
Johnson, A.E., Hebert, M.: Using spin images for efficient object recognition
  in cluttered {3D} scenes.
\newblock IEEE Transactions on pattern analysis and machine intelligence
  \textbf{21}(5), 433--449 (1999)

\bibitem{2DColour}
Jumb, V., Sohani, M., Shrivas, A.: Color image segmentation using {K-means}
  clustering and {Otsu’s} adaptive thresholding.
\newblock International Journal of Innovative Technology and Exploring
  Engineering (IJITEE) \textbf{3}(9), 72--76 (2014)

\bibitem{kaiser2018autonomous}
Kaiser, P., Asfour, T.: Autonomous detection and experimental validation of
  affordances.
\newblock IEEE Robotics and Automation Letters \textbf{3}(3), 1949--1956 (2018)

\bibitem{kang2019sampling}
Kang, G., Kim, Y.B., Lee, Y.H., Oh, H.S., You, W.S., Choi, H.R.: Sampling-based
  motion planning of manipulator with goal-oriented sampling.
\newblock Intelligent Service Robotics pp. 1--9 (2019)

\bibitem{kappler2018}
{Kappler}, D., {Meier}, F., {Issac}, J., {Mainprice}, J., {Cifuentes}, C.G.,
  {Wüthrich}, M., {Berenz}, V., {Schaal}, S., {Ratliff}, N., {Bohg}, J.:
  Real-time perception meets reactive motion generation.
\newblock IEEE Robotics and Automation Letters \textbf{3}(3), 1864--1871 (2018)

\bibitem{karaman2011globalnav}
Karaman, S., Frazzoli, E.: Sampling-based algorithms for optimal motion
  planning.
\newblock The International Journal of Robotics Research \textbf{30}(7),
  846--894 (2011).
\newblock \doi{10.1177/0278364911406761}.
\newblock \urlprefix\url{https://doi.org/10.1177/0278364911406761}

\bibitem{kasaei2019orthographicnet}
Kasaei, H.: {OrthographicNet}: A deep learning approach for {3D} object
  recognition in open-ended domains.
\newblock arXiv preprint arXiv:1902.03057  (2019)

\bibitem{kasaei2019look}
Kasaei, S.H.: Look further to recognize better: Learning shared topics and
  category-specific dictionaries for open-ended {3D} object recognition.
\newblock arXiv preprint arXiv:1907.12924  (2019)

\bibitem{KasaeiColorConstancy2020}
Kasaei, S.H., Ghorbani, M., Schilperoort, J., van~der Rest, W.: Investigating
  the importance of shape features, color constancy, color spaces and
  similarity measures in open-ended {3D} object recognition.
\newblock arXiv preprint arXiv:2002.03779  (2020)

\bibitem{kasaei2015adaptive}
Kasaei, S.H., Oliveira, M., Lim, G.H., Seabra~Lopes, L., Tom{\'e}, A.M.: An
  adaptive object perception system based on environment exploration and
  bayesian learning.
\newblock In: 2015 IEEE International Conference on Autonomous Robot Systems
  and Competitions, pp. 221--226. IEEE (2015)

\bibitem{kasaei2015interactive}
Kasaei, S.H., Oliveira, M., Lim, G.H., Seabra~Lopes, L., Tom{\'e}, A.M.:
  Interactive open-ended learning for {3D} object recognition: An approach and
  experiments.
\newblock Journal of Intelligent \& Robotic Systems \textbf{80}(3-4), 537--553
  (2015)

\bibitem{kasaei2018towards}
Kasaei, S.H., Oliveira, M., Lim, G.H., Seabra~Lopes, L., Tom{\'e}, A.M.:
  Towards lifelong assistive robotics: A tight coupling between object
  perception and manipulation.
\newblock Neurocomputing \textbf{291}, 151--166 (2018)

\bibitem{kasaei2016concurrent}
Kasaei, S.H., Seabra~Lopes, L., Tom{\'e}, A.M.: Concurrent {3D} object category
  learning and recognition based on topic modelling and human feedback.
\newblock In: 2016 International Conference on Autonomous Robot Systems and
  Competitions (ICARSC), pp. 329--334. IEEE (2016)

\bibitem{kasaei2019interactive}
Kasaei, S.H., Shafii, N., Seabra~Lopes, L., Tome, A.M.: Interactive open-ended
  object, affordance and grasp learning for robotic manipulation.
\newblock In: 2019 IEEE International Conference on Robotics and Automation
  (ICRA) (2019)

\bibitem{kasaei2018perceiving}
Kasaei, S.H., Sock, J., Seabra~Lopes, L., Tom{\'e}, A.M., Kim, T.K.:
  Perceiving, learning, and recognizing {3D} objects: An approach to cognitive
  service robots.
\newblock In: Thirty-Second AAAI Conference on Artificial Intelligence (2018)

\bibitem{kasaei2016hierarchical}
Kasaei, S.H., Tom{\'e}, A.M., Seabra~Lopes, L.: Hierarchical object
  representation for open-ended object category learning and recognition.
\newblock In: Advances in Neural Information Processing Systems, pp. 1948--1956
  (2016)

\bibitem{GOOD}
Kasaei, S.H., Tom{\'e}, A.M., Seabra~Lopes, L., Oliveira, M.: {GOOD}: A global
  orthographic object descriptor for {3D} object recognition and manipulation.
\newblock Pattern Recognition Letters \textbf{83}, 312--320 (2016)

\bibitem{Local-LDA}
Kasaei, S.H.M., Seabra~Lopes, L., Tom{\'e}, A.M.: {Local-LDA}: Open-ended
  learning of latent topics for {3D} object recognition.
\newblock IEEE transactions on pattern analysis and machine intelligence (PAMI)
   (2019)

\bibitem{cloudReview1}
Kehoe, B., Patil, S., Abbeel, P., Goldberg, K.: A survey of research on cloud
  robotics and automation.
\newblock IEEE Transactions on automation science and engineering
  \textbf{12}(2), 398--409 (2015)

\bibitem{kemker2018measuring}
Kemker, R., McClure, M., Abitino, A., Hayes, T.L., Kanan, C.: Measuring
  catastrophic forgetting in neural networks.
\newblock In: Thirty-second AAAI conference on artificial intelligence (2018)

\bibitem{kertesz2018common}
Kert{\'e}sz, C., Turunen, M.: Common sounds in bedrooms (csibe) corpora for
  sound event recognition of domestic robots.
\newblock Intelligent Service Robotics \textbf{11}(4), 335--346 (2018)

\bibitem{tasteAndVision}
Kiani, S., Minaei, S., Ghasemi-Varnamkhasti, M.: Fusion of artificial senses as
  a robust approach to food quality assessment.
\newblock Journal of Food Engineering \textbf{171}, 230--239 (2016)

\bibitem{kim2019integration}
Kim, B.W., Park, Y., Suh, I.H.: Integration of top-down and bottom-up visual
  processing using a recurrent convolutional--deconvolutional neural network
  for semantic segmentation.
\newblock Intelligent Service Robotics pp. 1--11 (2019)

\bibitem{kirkpatrick2017overcoming}
Kirkpatrick, J., Pascanu, R., Rabinowitz, N., Veness, J., Desjardins, G., Rusu,
  A.A., Milan, K., Quan, J., Ramalho, T., Grabska-Barwinska, A., et~al.:
  Overcoming catastrophic forgetting in neural networks.
\newblock Proceedings of the national academy of sciences \textbf{114}(13),
  3521--3526 (2017)

\bibitem{kopicki2016one}
Kopicki, M., Detry, R., Adjigble, M., Stolkin, R., Leonardis, A., Wyatt, J.L.:
  One-shot learning and generation of dexterous grasps for novel objects.
\newblock The International Journal of Robotics Research \textbf{35}(8),
  959--976 (2016)

\bibitem{kopicki2019learning}
Kopicki, M.S., Belter, D., Wyatt, J.L.: Learning better generative models for
  dexterous, single-view grasping of novel objects.
\newblock The International Journal of Robotics Research \textbf{38}(10-11),
  1246--1267 (2019)

\bibitem{Kragic2003}
Kragic, D., Christensen, H.: Robust visual servoing.
\newblock The International Journal of Robotics Research \textbf{22}(10-11),
  923--939 (2003).
\newblock 923

\bibitem{krawczyk2015one}
Krawczyk, B., Wo{\'z}niak, M.: One-class classifiers with incremental learning
  and forgetting for data streams with concept drift.
\newblock Soft Computing \textbf{19}(12), 3387--3400 (2015)

\bibitem{krizhevsky2012imagenet}
Krizhevsky, A., Sutskever, I., Hinton, G.E.: Imagenet classification with deep
  convolutional neural networks.
\newblock In: Advances in neural information processing systems, pp. 1097--1105
  (2012)

\bibitem{langley2009cognitive}
Langley, P., Laird, J.E., Rogers, S.: Cognitive architectures: Research issues
  and challenges.
\newblock Cognitive Systems Research \textbf{10}(2), 141--160 (2009)

\bibitem{lebedev2002insights}
Lebedev, M.A., Wise, S.P.: Insights into seeing and grasping: distinguishing
  the neural correlates of perception and action.
\newblock Behavioral and cognitive neuroscience reviews \textbf{1}(2), 108--129
  (2002)

\bibitem{lenz2015deep}
Lenz, I., Lee, H., Saxena, A.: Deep learning for detecting robotic grasps.
\newblock The International Journal of Robotics Research \textbf{34}(4-5),
  705--724 (2015)

\bibitem{Armen}
Leroux, C., Lebec, O., Ghezala, M.B., Mezouar, Y., Devillers, L., Chastagnol,
  C., Martin, J.C., Leynaert, V., Fattal, C.: Armen: Assistive robotics to
  maintain elderly people in natural environment.
\newblock IRBM \textbf{34}(2), 101--107 (2013)

\bibitem{li2013sketch}
Li, B., Lu, Y., Johan, H.: Sketch-based {3D} model retrieval by viewpoint
  entropy-based adaptive view clustering.
\newblock In: Proceedings of the Sixth Eurographics Workshop on {3D} Object
  Retrieval, pp. 49--56. Eurographics Association (2013)

\bibitem{li2018pointcnn}
Li, Y., Bu, R., Sun, M., Wu, W., Di, X., Chen, B.: {PointCNN}: Convolution on
  x-transformed points.
\newblock In: Advances in neural information processing systems, pp. 820--830
  (2018)

\bibitem{instanceSegmentation}
Liang, X., Lin, L., Wei, Y., Shen, X., Yang, J., Yan, S.: Proposal-free network
  for instance-level object segmentation.
\newblock IEEE transactions on pattern analysis and machine intelligence
  \textbf{40}(12), 2978--2991 (2017)

\bibitem{liang2016intelligent}
Liang, Y.h., Cai, C.: Intelligent collision avoidance based on two-dimensional
  risk model.
\newblock Journal of Algorithms \& Computational Technology \textbf{10}(3),
  131--141 (2016)

\bibitem{contextMicrosoft}
Lin, T.Y., Maire, M., Belongie, S., Hays, J., Perona, P., Ramanan, D.,
  Doll{\'a}r, P., Zitnick, C.L.: Microsoft {COCO}: Common objects in context.
\newblock In: European conference on computer vision, pp. 740--755. Springer
  (2014)

\bibitem{long2015fully}
Long, J., Shelhamer, E., Darrell, T.: Fully convolutional networks for semantic
  segmentation.
\newblock In: Proceedings of the IEEE conference on computer vision and pattern
  recognition, pp. 3431--3440 (2015)

\bibitem{luddecke2019context}
L{\"u}ddecke, T., Kulvicius, T., W{\"o}rg{\"o}tter, F.: Context-based
  affordance segmentation from {2D} images for robot actions.
\newblock Robotics and Autonomous Systems  (2019)

\bibitem{luo2017robotic}
Luo, S., Bimbo, J., Dahiya, R., Liu, H.: Robotic tactile perception of object
  properties: A review.
\newblock Mechatronics \textbf{48}, 54--67 (2017)

\bibitem{Sha2020}
Luo, S., Kasaei, H., Schomaker, L.: Accelerating reinforcement learning for
  reaching using continuous curriculum learning.
\newblock arXiv preprint arXiv:2002.02697  (2020)

\bibitem{mahler2017dex}
Mahler, J., Liang, J., Niyaz, S., Laskey, M., Doan, R., Liu, X., Ojea, J.A.,
  Goldberg, K.: Dex-net 2.0: Deep learning to plan robust grasps with synthetic
  point clouds and analytic grasp metrics.
\newblock arXiv preprint arXiv:1703.09312  (2017)

\bibitem{mahlerdex3}
Mahler, J., Matl, M., Liu, X., Li, A., Gealy, D., Goldberg, K.: {Dex-Net} 3.0:
  Computing robust robot vacuum suction grasp targets in point clouds using a
  new analytic model and deep learning.
\newblock arXiv preprint arXiv:1709.06670  (2017)

\bibitem{mahler2019learning}
Mahler, J., Matl, M., Satish, V., Danielczuk, M., DeRose, B., McKinley, S.,
  Goldberg, K.: Learning ambidextrous robot grasping policies.
\newblock Science Robotics \textbf{4}(26) (2019)

\bibitem{mahler2016dex}
Mahler, J., Pokorny, F.T., Hou, B., Roderick, M., Laskey, M., Aubry, M.,
  Kohlhoff, K., Kr{\"o}ger, T., Kuffner, J., Goldberg, K.: {Dex-Net} 1.0: A
  cloud-based network of {3D} objects for robust grasp planning using a
  multi-armed bandit model with correlated rewards.
\newblock In: 2016 IEEE international conference on robotics and automation
  (ICRA), pp. 1957--1964. IEEE (2016)

\bibitem{mauro2014unified}
Mauro, M., Riemenschneider, H., Signoroni, A., Leonardi, R., Van~Gool, L.: A
  unified framework for content-aware view selection and planning through view
  importance.
\newblock Proceedings BMVC 2014 pp. 1--11 (2014)

\bibitem{robocup2018}
Memmesheimer, R., Mykhalchyshyna, I., Seib, V., Evers, T., Paulus, D.:
  homer@UniKoblenz: Winning Team of the RoboCup@Home Open Platform League 2018,
  pp. 512--523.
\newblock Springer International Publishing (2019).
\newblock \doi{10.1007/978-3-030-27544-0_42}

\bibitem{metta2010icub}
Metta, G., Natale, L., Nori, F., Sandini, G., Vernon, D., Fadiga, L.,
  Von~Hofsten, C., Rosander, K., Lopes, M., Santos-Victor, J., et~al.: The
  {iCub} humanoid robot: An open-systems platform for research in cognitive
  development.
\newblock Neural Networks \textbf{23}(8-9), 1125--1134 (2010)

\bibitem{miller2004graspit}
{Miller}, A.T., {Allen}, P.K.: {GraspIt!} a versatile simulator for robotic
  grasping.
\newblock IEEE Robotics Automation Magazine \textbf{11}(4), 110--122 (2004)

\bibitem{mirrazavi2018unified}
Mirrazavi~Salehian, S.S., Figueroa, N., Billard, A.: A unified framework for
  coordinated multi-arm motion planning.
\newblock The International Journal of Robotics Research \textbf{37}(10),
  1205--1232 (2018)

\bibitem{mo2019partnet}
Mo, K., Zhu, S., Chang, A.X., Yi, L., Tripathi, S., Guibas, L.J., Su, H.:
  {PartNet}: A large-scale benchmark for fine-grained and hierarchical
  part-level {3D} object understanding.
\newblock In: Proceedings of the IEEE Conference on Computer Vision and Pattern
  Recognition, pp. 909--918 (2019)

\bibitem{moll2017randomized}
Moll, M., Kavraki, L., Rosell, J., et~al.: Randomized physics-based motion
  planning for grasping in cluttered and uncertain environments.
\newblock IEEE Robotics and Automation Letters \textbf{3}(2), 712--719 (2017)

\bibitem{moriello2018manipulating}
Moriello, L., Biagiotti, L., Melchiorri, C., Paoli, A.: Manipulating liquids
  with robots: A sloshing-free solution.
\newblock Control Engineering Practice \textbf{78}, 129--141 (2018)

\bibitem{morrison2018closing}
Morrison, D., Corke, P., Leitner, J.: {Closing the Loop for Robotic Grasping: A
  Real-time, Generative Grasp Synthesis Approach}.
\newblock In: Proc.\ of Robotics: Science and Systems (RSS) (2018)

\bibitem{GGCNN}
Morrison, D., Corke, P., Leitner, J.: Closing the loop for robotic grasping: A
  real-time, generative grasp synthesis approach.
\newblock In: International Conference on Robotics: Science and Systems (RSS)
  (2018)

\bibitem{morrison2019learning}
Morrison, D., Corke, P., Leitner, J.: Learning robust, real-time, reactive
  robotic grasping.
\newblock The International Journal of Robotics Research pp. 183--201 (2019)

\bibitem{mottaghi2014role}
Mottaghi, R., Chen, X., Liu, X., Cho, N.G., Lee, S.W., Fidler, S., Urtasun, R.,
  Yuille, A.: The role of context for object detection and semantic
  segmentation in the wild.
\newblock In: Proceedings of the IEEE Conference on Computer Vision and Pattern
  Recognition, pp. 891--898 (2014)

\bibitem{mukadam2018continuous}
Mukadam, M., Dong, J., Yan, X., Dellaert, F., Boots, B.: Continuous-time
  gaussian process motion planning via probabilistic inference.
\newblock The International Journal of Robotics Research \textbf{37}(11),
  1319--1340 (2018)

\bibitem{grasp6D}
Murali, A., Mousavian, A., Eppner, C., Paxton, C., Fox, D.: {6-DOF} grasping
  for target-driven object manipulation in clutter.
\newblock In: 2020 IEEE International Conference on Robotics and Automation
  (ICRA), pp. 1--8. IEEE (2020)

\bibitem{myers2015affordance}
Myers, A., Teo, C.L., Ferm{\"u}ller, C., Aloimonos, Y.: Affordance detection of
  tool parts from geometric features.
\newblock In: 2015 IEEE International Conference on Robotics and Automation
  (ICRA), pp. 1374--1381. IEEE (2015)

\bibitem{tasteFabrication}
Nag, A., Mukhopadhyay, S.C.: Fabrication and implementation of printed sensors
  for taste sensing applications.
\newblock Sensors and Actuators A: Physical \textbf{269}, 53--61 (2018)

\bibitem{smellTasteStandards}
Nagle, H.T., Schiffman, S.S.: Electronic taste and smell: the case for
  performance standards [point of view].
\newblock Proceedings of the IEEE \textbf{106}(9), 1471--1478 (2018)

\bibitem{cloudPerception2}
Niemueller, T., Schiffer, S., Lakemeyer, G., Rezapour-Lakani, S.: Life-long
  learning perception using cloud database technology.
\newblock In: Proc. IROS Workshop on Cloud Robotics. Citeseer (2013)

\bibitem{oliveira2018efficient}
Oliveira, G.L., Bollen, C., Burgard, W., Brox, T.: Efficient and robust deep
  networks for semantic segmentation.
\newblock The International Journal of Robotics Research \textbf{37}(4-5),
  472--491 (2018)

\bibitem{oliveira2014perceptual}
Oliveira, M., Lim, G.H., Seabra~Lopes, L., Kasaei, S.H., Tom{\'e}, A.M.,
  Chauhan, A.: A perceptual memory system for grounding semantic
  representations in intelligent service robots.
\newblock In: 2014 IEEE/RSJ International Conference on Intelligent Robots and
  Systems, pp. 2216--2223. IEEE (2014)

\bibitem{bagOfWords}
Oliveira, M., Seabra~Lopes, L., Lim, G.H., Kasaei, S.H., Sappa, A.D., Tom{\'e},
  A.M.: Concurrent learning of visual codebooks and object categories in
  open-ended domains.
\newblock In: 2015 IEEE/RSJ International Conference on Intelligent Robots and
  Systems (IROS), pp. 2488--2495. IEEE (2015)

\bibitem{oliveira20163d}
Oliveira, M., Seabra~Lopes, L., Lim, G.H., Kasaei, S.H., Tom{\'e}, A.M.,
  Chauhan, A.: {3D} object perception and perceptual learning in the {RACE}
  project.
\newblock Robotics and Autonomous Systems \textbf{75}, 614--626 (2016)

\bibitem{andrychowicz2020dextrous}
OM, A., B, B., M, C., R, J., B, M., J, P., A, P., M, P., G, P.: Learning
  dexterous in-hand manipulation.
\newblock International Journal of Robotics Research \textbf{39}(1), 3--20
  (2020)

\bibitem{oreshkin2018tadam}
Oreshkin, B., L{\'o}pez, P.R., Lacoste, A.: {TADAM}: Task dependent adaptive
  metric for improved few-shot learning.
\newblock In: Advances in Neural Information Processing Systems, pp. 721--731
  (2018)

\bibitem{pages2016tiago}
Pages, J., Marchionni, L., Ferro, F.: Tiago: the modular robot that adapts to
  different research needs.
\newblock In: International workshop on robot modularity, IROS (2016)

\bibitem{pan2012collision}
Pan, J., Zhang, L., Manocha, D.: Collision-free and smooth trajectory
  computation in cluttered environments.
\newblock The International Journal of Robotics Research \textbf{31}(10),
  1155--1175 (2012)

\bibitem{parisi2019continual}
Parisi, G.I., Kemker, R., Part, J.L., Kanan, C., Wermter, S.: Continual
  lifelong learning with neural networks: A review.
\newblock Neural Networks \textbf{113}, 54--71 (2019)

\bibitem{piazza2019century}
Piazza, C., Grioli, G., Catalano, M., Bicchi, A.: A century of robotic hands.
\newblock Annual Review of Control, Robotics, and Autonomous Systems
  \textbf{2}, 1--32 (2019)

\bibitem{nextBestSmell}
Polvara, R., Trabattoni, M., Kucner, T.P., Schaffernicht, E., Amigoni, F.,
  Lilienthal, A.J.: A next-best-smell approach for remote gas detection with a
  mobile robot.
\newblock arXiv preprint arXiv:1801.06819  (2018)

\bibitem{qi2017pointnet}
Qi, C.R., Su, H., Mo, K., Guibas, L.J.: {PointNet:} deep learning on point sets
  for {3D} classification and segmentation.
\newblock In: Proceedings of the IEEE conference on computer vision and pattern
  recognition, pp. 652--660 (2017)

\bibitem{qi2017pointnet++}
Qi, C.R., Yi, L., Su, H., Guibas, L.J.: {PointNet++}: Deep hierarchical feature
  learning on point sets in a metric space.
\newblock In: Advances in neural information processing systems, pp. 5099--5108
  (2017)

\bibitem{qian2020grasp}
Qian, K., Jing, X., Duan, Y., Zhou, B., Fang, F., Xia, J., Ma, X.: Grasp pose
  detection with affordance-based task constraint learning in single-view point
  clouds.
\newblock JOURNAL OF INTELLIGENT \& ROBOTIC SYSTEMS  (2020)

\bibitem{qin2019s4g}
Qin, Y., Chen, R., Zhu, H., Song, M., Xu, J., Su, H.: {S4G}: Amodal single-view
  single-shot {SE(3)} grasp detection in cluttered scenes.
\newblock arXiv preprint arXiv:1910.14218  (2019)

\bibitem{EQuality}
Qiu, S., Wang, J.: Application of sensory evaluation, hs-spme gc-ms, e-nose,
  and e-tongue for quality detection in citrus fruits.
\newblock Journal of food science \textbf{80}(10), S2296--S2304 (2015)

\bibitem{qureshi2019motion}
Qureshi, A.H., Miao, Y., Simeonov, A., Yip, M.C.: Motion planning networks:
  Bridging the gap between learning-based and classical motion planners.
\newblock arXiv preprint arXiv:1907.06013  (2019)

\bibitem{qureshi2019motionICRA}
Qureshi, A.H., Simeonov, A., Bency, M.J., Yip, M.C.: Motion planning networks.
\newblock In: 2019 International Conference on Robotics and Automation (ICRA),
  pp. 2118--2124. IEEE (2019)

\bibitem{contextBasedRecognition}
Rabinovich, A., Vedaldi, A., Galleguillos, C., Wiewiora, E., Belongie, S.J.:
  Objects in context.
\newblock In: ICCV, vol.~1, pp. 1--8. Citeseer (2007)

\bibitem{rakita2018relaxedik}
Rakita, D., Mutlu, B., Gleicher, M.: {RelaxedIK}: Real-time synthesis of
  accurate and feasible robot arm motion.
\newblock In: Robotics: Science and Systems (2018)

\bibitem{redmon2016you}
Redmon, J., Divvala, S., Girshick, R., Farhadi, A.: You only look once:
  Unified, real-time object detection.
\newblock In: Proceedings of the IEEE conference on computer vision and pattern
  recognition, pp. 779--788 (2016)

\bibitem{fasterRCNN}
Ren, S., He, K., Girshick, R., Sun, J.: Faster r-cnn: Towards real-time object
  detection with region proposal networks.
\newblock In: Advances in neural information processing systems, pp. 91--99
  (2015)

\bibitem{ristin2014incremental}
Ristin, M., Guillaumin, M., Gall, J., Van~Gool, L.: Incremental learning of
  {NCM} forests for large-scale image classification.
\newblock In: Proceedings of the IEEE conference on computer vision and pattern
  recognition, pp. 3654--3661 (2014)

\bibitem{FPFH}
Rusu, R.B., Blodow, N., Beetz, M.: Fast point feature histograms ({FPFH}) for
  {3D} registration.
\newblock In: 2009 IEEE International Conference on Robotics and Automation,
  pp. 3212--3217. IEEE (2009)

\bibitem{VFH}
Rusu, R.B., Bradski, G., Thibaux, R., Hsu, J.: Fast {3D} recognition and pose
  using the viewpoint feature histogram.
\newblock In: 2010 IEEE/RSJ International Conference on Intelligent Robots and
  Systems, pp. 2155--2162. IEEE (2010)

\bibitem{PFH}
Rusu, R.B., Marton, Z.C., Blodow, N., Beetz, M.: Learning informative point
  classes for the acquisition of object model maps.
\newblock In: 2008 10th International Conference on Control, Automation,
  Robotics and Vision, pp. 643--650. IEEE (2008)

\bibitem{cloudReview2}
Saha, O., Dasgupta, P.: A comprehensive survey of recent trends in cloud
  robotics architectures and applications.
\newblock Robotics \textbf{7}(3), 47 (2018)

\bibitem{sahbani2012overview}
Sahbani, A., El-Khoury, S., Bidaud, P.: An overview of {3D} object grasp
  synthesis algorithms.
\newblock Robotics and Autonomous Systems \textbf{60}(3), 326--336 (2012)

\bibitem{ClearGrasp}
Sajjan, S.S., Moore, M., Pan, M., Nagaraja, G., Lee, J., Zeng, A., Song, S.:
  {ClearGrasp}:{3D} shape estimation of transparent objects for manipulation.
\newblock In: 2020 IEEE International Conference on Robotics and Automation
  (ICRA) (2020)

\bibitem{shafii2016learning}
Shafii, N., Kasaei, S.H., Seabra~Lopes, L.: Learning to grasp familiar objects
  using object view recognition and template matching.
\newblock In: 2016 IEEE/RSJ International Conference on Intelligent Robots and
  Systems (IROS), pp. 2895--2900. IEEE (2016)

\bibitem{IoT}
Simoens, P., Dragone, M., Saffiotti, A.: The internet of robotic things: A
  review of concept, added value and applications.
\newblock International Journal of Advanced Robotic Systems \textbf{15}(1),
  1729881418759424 (2018)

\bibitem{singh2019neural}
Singh, N.H., Thongam, K.: Neural network-based approaches for mobile robot
  navigation in static and moving obstacles environments.
\newblock Intelligent Service Robotics \textbf{12}(1), 55--67 (2019)

\bibitem{skovcaj2016integrated}
Sko{\v{c}}aj, D., Vre{\v{c}}ko, A., Mahni{\v{c}}, M., Jan{\'\i}{\v{c}}ek, M.,
  Kruijff, G.J.M., Hanheide, M., Hawes, N., Wyatt, J.L., Keller, T., Zhou, K.,
  et~al.: An integrated system for interactive continuous learning of
  categorical knowledge.
\newblock Journal of Experimental \& Theoretical Artificial Intelligence
  \textbf{28}(5), 823--848 (2016)

\bibitem{sock2017multi}
Sock, J., Kasaei, S.H., Seabra~Lopes, L., Kim, T.K.: Multi-view {6D} object
  pose estimation and camera motion planning using rgbd images.
\newblock In: Proceedings of the IEEE International Conference on Computer
  Vision Workshops, pp. 2228--2235 (2017)

\bibitem{spasova2018challenges}
Spasova, S., Baeten, R., Coster, S., Ghailani, D., Pe{\~n}a-Casas, R.,
  Vanhercke, B.: Challenges in long-term care in europe.
\newblock A study of national policies, European Social Policy Network (ESPN),
  Brussels: European Commission  (2018)

\bibitem{Busboy}
Srinivasa, S., Ferguson, D., Vandeweghe, J.M., Diankov, R., Berenson, D.,
  Helfrich, C., Strasdat, K.: The robotic busboy: Steps towards developing a
  mobile robotic home assistant.
\newblock In: Proceedings of International Conference on Intelligent Autonomous
  Systems (2008)

\bibitem{HERB}
Srinivasa, S.S., Ferguson, D., Helfrich, C.J., Berenson, D., Collet, A.,
  Diankov, R., Gallagher, G., Hollinger, G., Kuffner, J., Weghe, M.V.: {HERB}:
  a home exploring robotic butler.
\newblock Autonomous Robots \textbf{28}(1), 5 (2010)

\bibitem{stein1993merging}
Stein, B.E., Meredith, M.A.: The merging of the senses.
\newblock The MIT Press (1993)

\bibitem{stilman2007manipulation}
Stilman, M., Schamburek, J.U., Kuffner, J., Asfour, T.: Manipulation planning
  among movable obstacles.
\newblock In: Proceedings 2007 IEEE international conference on robotics and
  automation, pp. 3327--3332. IEEE (2007)

\bibitem{sun2010learning}
Sun, J., Moore, J.L., Bobick, A., Rehg, J.M.: Learning visual object categories
  for robot affordance prediction.
\newblock The International Journal of Robotics Research \textbf{29}(2-3),
  174--197 (2010)

\bibitem{sundaralingam2019relaxed}
Sundaralingam, B., Hermans, T.: Relaxed-rigidity constraints: kinematic
  trajectory optimization and collision avoidance for in-grasp manipulation.
\newblock Autonomous Robots \textbf{43}(2), 469--483 (2019)

\bibitem{szegedy2013deep}
Szegedy, C., Toshev, A., Erhan, D.: Deep neural networks for object detection.
\newblock In: Advances in neural information processing systems, pp. 2553--2561
  (2013)

\bibitem{foodQuality}
Tan, J., Xu, J.: Applications of electronic nose (e-nose) and electronic tongue
  (e-tongue) in food quality-related properties determination: A review.
\newblock Artificial Intelligence in Agriculture  (2020)

\bibitem{FCOS}
Tian, Z., Shen, C., Chen, H., He, T.: Fcos: Fully convolutional one-stage
  object detection.
\newblock In: Proceedings of the IEEE international conference on computer
  vision, pp. 9627--9636 (2019)

\bibitem{toussaint2018differentiable}
Toussaint, M., Allen, K.R., Smith, K.A., Tenenbaum, J.B.: Differentiable
  physics and stable modes for tool-use and manipulation planning.
\newblock In: Robotics: Science and Systems, vol.~2 (2018)

\bibitem{truong2017socially}
Truong, X.T., Yoong, V.N., Ngo, T.D.: Socially aware robot navigation system in
  human interactive environments.
\newblock Intelligent Service Robotics \textbf{10}(4), 287--295 (2017)

\bibitem{Walk-man}
Tsagarakis, N.G., Caldwell, D.G., Negrello, F., Choi, W., Baccelliere, L., Loc,
  V.G., Noorden, J., Muratore, L., Margan, A., Cardellino, A., et~al.:
  {WALK-MAN}: A high-performance humanoid platform for realistic environments.
\newblock Journal of Field Robotics \textbf{34}(7), 1225--1259 (2017)

\bibitem{tsarouchi2016human}
Tsarouchi, P., Makris, S., Chryssolouris, G.: Human--robot interaction review
  and challenges on task planning and programming.
\newblock International Journal of Computer Integrated Manufacturing
  \textbf{29}(8), 916--931 (2016)

\bibitem{tschannen2018recent}
Tschannen, M., Bachem, O., Lucic, M.: Recent advances in autoencoder-based
  representation learning.
\newblock arXiv preprint arXiv:1812.05069  (2018)

\bibitem{ullrich2017selecting}
Ullrich, M., Ali, H., Durner, M., M{\'a}rton, Z.C., Triebel, R.: Selecting
  {CNN} features for online learning of {3D} objects.
\newblock In: 2017 IEEE/RSJ International Conference on Intelligent Robots and
  Systems (IROS), pp. 5086--5091. IEEE (2017)

\bibitem{van2009path}
Van Den~Berg, J., Stilman, M., Kuffner, J., Lin, M., Manocha, D.: Path planning
  among movable obstacles: a probabilistically complete approach.
\newblock In: Algorithmic Foundation of Robotics VIII, pp. 599--614. Springer
  (2009)

\bibitem{van2014probabilistic}
Van~Hoof, H., Kroemer, O., Peters, J.: Probabilistic segmentation and targeted
  exploration of objects in cluttered environments.
\newblock IEEE Transactions on Robotics \textbf{30}(5), 1198--1209 (2014)

\bibitem{regionGrowing2D}
Verma, O.P., Hanmandlu, M., Susan, S., Kulkarni, M., Jain, P.K.: A simple
  single seeded region growing algorithm for color image segmentation using
  adaptive thresholding.
\newblock In: 2011 International Conference on Communication Systems and
  Network Technologies, pp. 500--503. IEEE (2011)

\bibitem{vezzani2017novel}
Vezzani, G., Regoli, M., Pattacini, U., Natale, L.: A novel pipeline for
  bi-manual handover task.
\newblock Advanced Robotics \textbf{31}(23-24), 1267--1280 (2017)

\bibitem{bioSmeller}
Villarreal, B.L., Gordillo, J.: Bioinspired smell sensor: nostril model and
  design.
\newblock IEEE/ASME Transactions on Mechatronics \textbf{21}(2), 912--921
  (2015)

\bibitem{wang2019panet}
Wang, K., Liew, J.H., Zou, Y., Zhou, D., Feng, J.: Panet: Few-shot image
  semantic segmentation with prototype alignment.
\newblock In: Proceedings of the IEEE International Conference on Computer
  Vision, pp. 9197--9206 (2019)

\bibitem{SOLOv1}
Wang, X., Kong, T., Shen, C., Jiang, Y., Li, L.: Solo: Segmenting objects by
  locations.
\newblock arXiv preprint arXiv:1912.04488  (2019)

\bibitem{SOLOv2}
Wang, X., Zhang, R., Kong, T., Li, L., Shen, C.: Solov2: Dynamic, faster and
  stronger.
\newblock arXiv preprint arXiv:2003.10152  (2020)

\bibitem{wilfong1991motion}
Wilfong, G.: Motion planning in the presence of movable obstacles.
\newblock Annals of Mathematics and Artificial Intelligence \textbf{3}(1),
  131--150 (1991)

\bibitem{wise2016fetch}
Wise, M., Ferguson, M., King, D., Diehr, E., Dymesich, D.: Fetch and freight:
  Standard platforms for service robot applications.
\newblock In: Workshop on autonomous mobile service robots (2016)

\bibitem{ESF}
Wohlkinger, W., Vincze, M.: Ensemble of shape functions for {3D} object
  classification.
\newblock In: 2011 IEEE international conference on robotics and biomimetics,
  pp. 2987--2992. IEEE (2011)

\bibitem{wood2012review}
Wood, R., Baxter, P., Belpaeme, T.: A review of long-term memory in natural and
  synthetic systems.
\newblock Adaptive Behavior \textbf{20}(2), 81--103 (2012)

\bibitem{wu20153d}
Wu, Z., Song, S., Khosla, A., Yu, F., Zhang, L., Tang, X., Xiao, J.: {3D}
  {ShapeNets}: A deep representation for volumetric shapes.
\newblock In: Proceedings of the IEEE conference on computer vision and pattern
  recognition, pp. 1912--1920 (2015)

\bibitem{TasteAndVision2}
Xu, M., Wang, J., Zhu, L.: The qualitative and quantitative assessment of tea
  quality based on e-nose, e-tongue and e-eye combined with chemometrics.
\newblock Food chemistry \textbf{289}, 482--489 (2019)

\bibitem{xu2018spidercnn}
Xu, Y., Fan, T., Xu, M., Zeng, L., Qiao, Y.: {SpiderCNN}: Deep learning on
  point sets with parameterized convolutional filters.
\newblock In: Proceedings of the European Conference on Computer Vision (ECCV),
  pp. 87--102 (2018)

\bibitem{yasuda2020autonomous}
Yasuda, Y.D., Martins, L.E.G., Cappabianco, F.A.: Autonomous visual navigation
  for mobile robots: A systematic literature review.
\newblock ACM Computing Surveys (CSUR) \textbf{53}(1), 1--34 (2020)

\bibitem{behaviourBasedSmeller}
Yeon, A., Visvanathan, R., Mamduh, S., Kamarudin, K., Kamarudin, L., Zakaria,
  A.: Implementation of behaviour based robot with sense of smell and sight.
\newblock Procedia Computer Science \textbf{76}, 119--125 (2015)

\bibitem{yervilla2019optimal}
Yervilla-Herrera, H., Vasquez-Gomez, J.I., Murrieta-Cid, R., Becerra, I.,
  Sucar, L.E.: Optimal motion planning and stopping test for {3-D} object
  reconstruction.
\newblock Intelligent Service Robotics \textbf{12}(1), 103--123 (2019)

\bibitem{zeng2018robotic}
Zeng, A., Song, S., Yu, K.T., Donlon, E., Hogan, F.R., Bauza, M., Ma, D.,
  Taylor, O., Liu, M., Romo, E., et~al.: Robotic pick-and-place of novel
  objects in clutter with multi-affordance grasping and cross-domain image
  matching.
\newblock In: 2018 IEEE International Conference on Robotics and Automation
  (ICRA), pp. 1--8. IEEE (2018)

\bibitem{zeng2013mobile}
Zeng, L., Bone, G.M.: Mobile robot collision avoidance in human environments.
\newblock International Journal of Advanced Robotic Systems \textbf{10}(1), 41
  (2013)

\bibitem{colourSegmentationGPoint}
Zhan, Q., Liang, Y., Xiao, Y.: Color-based segmentation of point clouds.
\newblock Laser scanning \textbf{38}(3), 155--161 (2009)

\bibitem{zhang2016weakly}
Zhang, Y., Wei, X.S., Wu, J., Cai, J., Lu, J., Nguyen, V.A., Do, M.N.: Weakly
  supervised fine-grained categorization with part-based image representation.
\newblock IEEE Transactions on Image Processing \textbf{25}(4), 1713--1725
  (2016)

\bibitem{zhao2019sound}
Zhao, H., Gan, C., Ma, W.C., Torralba, A.: The sound of motions.
\newblock In: Proceedings of the IEEE International Conference on Computer
  Vision, pp. 1735--1744 (2019)

\bibitem{zhao2018sound}
Zhao, H., Gan, C., Rouditchenko, A., Vondrick, C., McDermott, J., Torralba, A.:
  The sound of pixels.
\newblock In: Proceedings of the European Conference on Computer Vision (ECCV),
  pp. 570--586 (2018)

\bibitem{zhao2019compatible}
Zhao, L., Liu, Z., Chen, J., Cai, W., Wang, W., Zeng, L.: A compatible
  framework for rgb-d slam in dynamic scenes.
\newblock IEEE Access \textbf{7}, 75604--75614 (2019)

\bibitem{zhao20193d}
Zhao, Y., Birdal, T., Deng, H., Tombari, F.: {3D} point capsule networks.
\newblock In: Proceedings of the IEEE Conference on Computer Vision and Pattern
  Recognition, pp. 1009--1018 (2019)

\bibitem{zhao2019object}
Zhao, Z.Q., Zheng, P., Xu, S.t., Wu, X.: Object detection with deep learning: A
  review.
\newblock IEEE transactions on neural networks and learning systems  (2019)

\bibitem{centernet}
Zhou, X., Wang, D., Kr{\"a}henb{\"u}hl, P.: Objects as points.
\newblock arXiv preprint arXiv:1904.07850  (2019)

\end{thebibliography}

\end{document}